%% file: main.tex
\documentclass[acmsmall, screen]{acmart}
\usepackage{amsmath,amsfonts}
\usepackage{graphicx}
\usepackage{textcomp}
\usepackage{xcolor}
\usepackage[english]{babel}
\usepackage[figuresright]{rotating}
\usepackage[flushleft]{threeparttable}
\newcommand{\toolname}{STP}
\usepackage{multirow}
\usepackage{bbm}
\usepackage{array}
\usepackage{graphicx} %
\usepackage{subfigure} %
\usepackage{fancybox}

\usepackage{algorithm}
\usepackage{algpseudocode}
\usepackage{supertabular,booktabs}
\usepackage{soul}
\soulregister\cite7 %
\soulregister\citep7 %
\soulregister\citet7 %
\soulregister\ref7 %
\soulregister\pageref7 %

\usepackage[normalem]{ulem}
\newcommand{\revise}[1]{{\color{black}{#1}}}
\newcommand{\delete}[1]{}

\newcommand{\newrevise}[1]{{\color{black}{#1}}}
\newcommand{\newdelete}[1]{}

\usepackage{url}
\usepackage{hyperref}
\hypersetup{
    colorlinks=true,
    linkcolor=blue,
    filecolor=magenta,      
    urlcolor=cyan,
}

\usepackage{CJKutf8}

\AtBeginDocument{
  \providecommand\BibTeX{{
    \normalfont B\kern-0.5em{\scshape i\kern-0.25em b}\kern-0.8em\TeX}}}

\setcopyright{acmcopyright}
\copyrightyear{2023}
\acmYear{2023}

\acmJournal{TOSEM}
\acmVolume{0}
\acmNumber{0}
\acmArticle{1}
\acmMonth{0}

\begin{document}

\title{Machine Translation Testing via Syntactic Tree Pruning}

\author{Quanjun Zhang} 
\orcid{0000-0002-2495-3805}
\email{quanjun.zhang@smail.nju.edu.cn}
\affiliation{
  \institution{State Key Laboratory for Novel Software Technology, Nanjing University}
  \city{Nanjing}
  \state{Jiangsu}
  \country{China}
  \postcode{210093}
}

\author{Juan Zhai} 
\orcid{0000-0001-5017-8016}
\email{juan.zhai@rutgers.edu}
\affiliation{
  \institution{Manning College of Information \& Computer Sciences, University of Massachusetts
Amherst}
  \city{Amherst}
  \state{MA}
  \country{USA}
  \postcode{01003}
}

\author{Chunrong Fang} 
\orcid{0000-0002-9930-7111}
\email{fangchunrong@nju.edu.cn}
\authornote{\textbf{Chunrong Fang is the corresponding author.}}
\affiliation{
  \institution{State Key Laboratory for Novel Software Technology, Nanjing University}
  \city{Nanjing}
  \state{Jiangsu}
  \country{China}
  \postcode{210093}
}

\author{Jiawei Liu} 
\orcid{0000-0002-4930-9637}
\email{jw.liu@smail.nju.edu.cn}
\affiliation{
  \institution{State Key Laboratory for Novel Software Technology, Nanjing University}
  \city{Nanjing}
  \state{Jiangsu}
  \country{China}
  \postcode{210093}
}

\author{Weisong Sun} 
\orcid{0000-0001-9236-8264}
\email{weisongsun@smail.nju.edu.cn}
\affiliation{
  \institution{State Key Laboratory for Novel Software Technology, Nanjing University}
  \city{Nanjing}
  \state{Jiangsu}
  \country{China}
  \postcode{210093}
}

\author{Haichuan Hu}
\orcid{0009-0002-3007-488X}
\email{181250046@smail.nju.edu.cn}
\affiliation{
  \institution{State Key Laboratory for Novel Software Technology, Nanjing University}
  \city{Nanjing}
  \state{Jiangsu}
  \country{China}
  \postcode{210093}
}

\author{Qingyu Wang}
\orcid{0009-0003-1693-4166}
\email{wangqingyu2012@gmail.com}
\affiliation{
  \institution{State Key Laboratory for Novel Software Technology, Nanjing University}
  \city{Nanjing}
  \state{Jiangsu}
  \country{China}
  \postcode{210093}
}

\begin{abstract}
Machine translation systems have been widely adopted in our daily life, making life easier and more convenient. Unfortunately, erroneous translations may result in severe consequences, such as financial losses. 
This requires to improve the accuracy and the reliability of machine translation systems. 
However, it is challenging to test machine translation systems because of the complexity and intractability of the underlying neural models.
To tackle these challenges, we propose a novel metamorphic testing approach by syntactic tree pruning ({\toolname}) to validate machine translation systems.
Our key insight is that a pruned sentence should have similar crucial semantics compared with the original sentence.
Specifically, {\toolname} 
(1) proposes a core semantics-preserving pruning strategy by basic sentence \delete{structure }\revise{structures} and dependency relations on the level of syntactic tree representation; 
(2) generates source sentence pairs based on the metamorphic relation; 
(3) reports suspicious issues whose translations break the consistency property by a bag-of-words model.
We further evaluate {\toolname} on two state-of-the-art machine translation systems (i.e., Google Translate and Bing Microsoft Translator) with 1,200 source sentences as inputs.
The results show that {\toolname} \delete{can }accurately \delete{find }\revise{finds} 5,073 unique erroneous translations in Google Translate and 5,100 unique erroneous translations in Bing Microsoft Translator (400\% more than state-of-the-art techniques), with 64.5\% and 65.4\% precision, respectively.
The reported erroneous translations vary in types and more than 90\% of them \newdelete{cannot be}\newrevise{are not} found by state-of-the-art techniques.
There are 9,393 erroneous translations unique to {\toolname}, which is 711.9\% more than state-of-the-art techniques.
Moreover, {\toolname} is quite effective \delete{to detect}\revise{in detecting} translation errors for the original sentences with a recall reaching 74.0\%, improving state-of-the-art techniques by 55.1\% on average.

\end{abstract}



\maketitle

\begin{CJK*}{UTF8}{gbsn}
\section{Introduction}
Machine translation aims to translate source content into target languages automatically.
Due to the advent of neural machine translation models, the performance of machine translation has been improved significantly.
In particular, some advanced machine translation systems are approaching human-level performance in terms of quality score and human parity \cite{2016Wu, 2018Hassan}.
More and more people are getting used to employing machine translation systems in their daily lives, such as reading articles from other countries.
For example, Google Translate \cite{Google} \delete{attracted }\revise{attracts} more than 500 million users \revise{around the world} and \delete{translated   }\revise{translates} more than 100 billion words per day \cite{2016Barak}.

Similar to traditional systems, neural machine translation systems are not perfect and suffer from unreliable results sometimes.
For example, they can produce erroneous target outputs when source inputs are adversarially manipulated (e.g., uppercasing some characters or injecting grammatical noise in a sentence) \cite{2018Belinkov, 2018Ebrahimi}.
These adversarial inputs are usually syntactically wrong.
Besides, there exist many cases where machine translation systems may return erroneous translations for syntactical and semantical inputs (e.g., an industrial case by WeChat) \cite{2019Zheng}.

However, there are several challenges to test machine translation systems.
Firstly, testing traditional systems significantly differs from testing DNN-based systems in general due to the programming paradigm \cite{2018ZhangTensorFlow}.
In the traditional system, \revise{the} decision logic is manifested in source code.
In contrast, the output of DNN-based systems depends mainly on the millions of parameters, which are optimized through training.
Secondly, recent testing approaches for DNN-based systems mainly focus on models with a small number of possible outputs (e.g., image classifiers).
Rather, it is an intractable problem for machine translation to enumerate all possible outputs \cite{2018Ott}, making machine translation systems challenging to test.
Thirdly, most existing machine translation testing techniques \cite{2020Gupta, 2020He, 2020Sun} generate test cases (i.e., generated source sentence) by replacing one word in the input (i.e., original source sentence) based on some pre-trained language representation models (i.e., BERT and Spacy).
Thus, the testing performance is mainly limited by the maturity of the adopted language model \cite{2021He} \revise{and huge device resources are required in the deployment \cite{shi2022compressing}}.

To address the above challenging problems, we introduce syntactic tree pruning ({\toolname}) testing, a novel and general approach for evaluating machine translation systems.
The core idea of {\toolname} is inspired by the phenomenon that machine translation systems usually perform significantly better on the simple sentence than \revise{on} the complex sentence \cite{2020Sikka}.
Thus, people prefer to conduct source sentence simplification to get accurate translation results in practice \cite{2021-Machine-Translation-Tips}.
The key insight is that eliminating \delete{irrelevant  }contextual information from a source sentence should not influence the translation results of the trunk~\cite{1988Mann, niklaus2019transforming} 
\newdelete{To realize this concept, {\toolname} generates new source sentences by removing words or phrases from an original source sentence without losing core semantics and undermining the sentence validity \revise{based on the linguistic rhetorical structure theory \cite{1988Mann}}.}
\newrevise{To realize this concept, inspired by the linguistic rhetorical structure theory~ \cite{1988Mann} that specifies each sentence component as either a basic nucleus or a context satellite, {\toolname} generates new source sentences by removing words or phrases from an original source sentence without losing core semantics and undermining the sentence validity.}
In particular, {\toolname} first performs a novel semantics-preserving pruning strategy by extracting crucial semantics via the basic sentence structure and designing pruning operators via dependency relations.
Then the original and the newly generated source sentences are paired via the defined metamorphic relation.
Besides, a bag-of-words model is adopted to measure the consistency of the source sentence pair (i.e., original and generated source sentence), and a suspicious issue will be reported if the translation results have a significant difference in core semantics \revise{via a pre-defined threshold value}.

We conduct an empirical study to evaluate the effectiveness of {\toolname} on two state-of-the-art translators (i.e., Google Translate and Bing Microsoft Translator) with 1,200 English sentences from ten \revise{major} article categories as inputs \cite{Google, Microsoft}.
\revise{The experimental results show that} {\toolname} successfully reveals 5,073 and 5,100 erroneous unique translations in Google Translate and Bing Microsoft Translator with high precision (\revise{i.e.,} 64.5\% and 65.4\%), respectively.
The detected erroneous translations are 400\% more than state-of-the-art techniques under all dataset categories.
We also demonstrate that the precision can be further improved under a large threshold \revise{value} $t$.
For example, {\toolname} with $t=6$ achieves 80.9\% and 84.3\% precision for \delete{the }both translators, outperforming the most recent technique CAT by 19.8\% and 18.2\% with \revise{a} comparable amount of erroneous issues.
The types of reported erroneous translations are diversified, including under-translation, over-translation, word/phrase mistranslation, incorrect modification, and unclear logic. 
Compared with state-of-the-art techniques, {\toolname} \delete{can }\revise{is able to} report more erroneous translations with higher precision.
Due to its conceptual difference, {\toolname} reveals many erroneous translations that have not been found by existing techniques.
For example, there are 4,700 and 4,692 erroneous translations unique to {\toolname} in Google Translate and Bing Microsoft Translator, which are 227\% and 221\% more than the most recent technique CAT.
{\toolname} also achieves a higher recall than \delete{sate-of-the-art }\revise{state-of-the-art} techniques by 53.4\% and 56.8\% on average for original sentence errors when testing Google Translate and Bing Microsoft Translator, respectively.
Besides, {\toolname} spends 0.09 seconds in sentence generation and 0.02 seconds in erroneous translation detection, \delete{achieve }\revise{achieving} comparable efficiency to state-of-the-art techniques.
All the reported issues and source code have been released for further research.

The main contributions of this paper are as follows.
\begin{itemize}
  \item \textbf{Novel Technique.} We introduce a novel and widely applicable methodology, \textit{syntactic tree pruning} ({\toolname}), to validate machine translation systems by generating sentences with different syntactic structures.

  \item \textbf{Practical Implementation.} We describe a practical implementation that generates new sentences via the proposed core semantic-preserving pruning strategy, and detects erroneous translations by measuring the semantics consistency of the source sentence pair, which is created by the designed metamorphic relation.
  
  \item \textbf{Extensive Study.} We evaluate {\toolname} on 1,200 sentences against five state-of-the-art techniques for Google Translate and Bing Microsoft Translator.
  To the best of our knowledge, this is the largest empirical evaluation for machine translation testing.
  The results demonstrate that {\toolname} successfully finds 5,073 erroneous translations in Google translate and 5,100 in Bing Microsoft Translator with high precision, most of which cannot be found by state-of-the-art techniques.
  On average, the recall of original sentence errors \delete{can reach }\revise{reaches} 74\%, which is 55.1\% higher than that of state-of-the-art techniques.
  
  \item \textbf{Available Artifacts.} We release all experimental data (including the raw data, source code, and result analysis) for replication and future research on machine translation testing \cite{2021-Open-Datasets}.
\end{itemize}

\revise{
 The rest of this paper is organized as follows.
 Section \ref{sec:bg&mv} reviews some background information and presents a motivation example.
 Section \ref{sec:approach} introduces the proposed approach design.
 Section \ref{sec:exp} presents the research questions, and explains the details of the empirical study.
 Section \ref{sec:eva} provides the detailed results of the study and answers the research questions.
 Section \ref{sec:dis} presents some additional discussion,  and Section \ref{sec:threats} discloses the threats to validity of our experiments.
 Section \ref{sec:rw} discusses some related work.
 Section \ref{sec:conclusion} presents the conclusions and discusses future work.
}

\section{Background \& Motivation}
\label{sec:bg&mv}

\subsection{Basic Sentence Structure}
\label{sec:basic_sentence}

In linguistics, given a source sentence, every word serves a specific purpose within the structure \cite{1984Huddleston}.
Sentence structure can sometimes be quite complicated according to the grammar rules.
In syntax, four types of sentence structures are distinguished: simple sentences, compound sentences, complex sentences and compound-complex sentences \cite{2010Quirk}.
Each sentence is defined by the usage of independent and dependent clauses, conjunctions and subordinators.
The type of sentence is determined by how many clauses, or subject–verb groups are included in the sentence.

Although the sentences are diverse in types, a complex linguistic structure sentence can be converted into a set of simple sentences, with each of which presents a simpler and more regular structure that is easier to process by machine translation systems \cite{manning2014stanford}.
A simple sentence is a sentence that comprises exactly one independent clause, i.e., a group of words that has both a subject (S) and a verb (V), optionally an indirect or direct object (O), and complement (C).
In linguistic typology, most simple sentences are derived based on the five basic clause types: subject–verb (SV), subject–verb–object (SVO), subject–verb–complement (SVC), subject–verb–indirect object–direct object (SVOO), subject–object–complement (SVOC) \cite{1995Lyons}.
However, simple sentences may still include optional constituents that render them overly complex.
For instance, the adverbial “in Princeton” in the simple sentence ``Albert Einstein died in Princeton.'' specifies additional contextual information that can be left out without producing ill-formed output. 
Rather, the remaining clause ``Albert Einstein died.'' still carries semantically meaningful information.

The constituents that belong to the clause type are essential components of the corresponding simple sentence, all other constituents are optional and can be discarded without leading to an incoherent or semantically meaningless output.
In this paper, given a source sentence, we attempt to identify and simplify the sentence types, and then extract the basic clauses as the crucial semantics.

\subsection{Syntactic Tree Representation}

\begin{figure}[htbp]
\centering

	\subfigure[Constituency Structure] {
		\includegraphics[width=0.47\columnwidth]{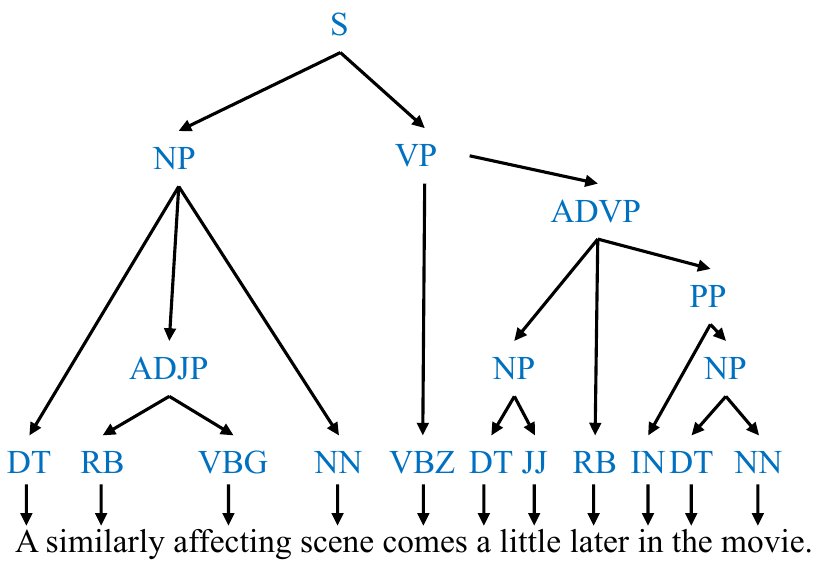}
	}
	\subfigure[Dependency Structure] {
		\includegraphics[width=0.47\columnwidth]{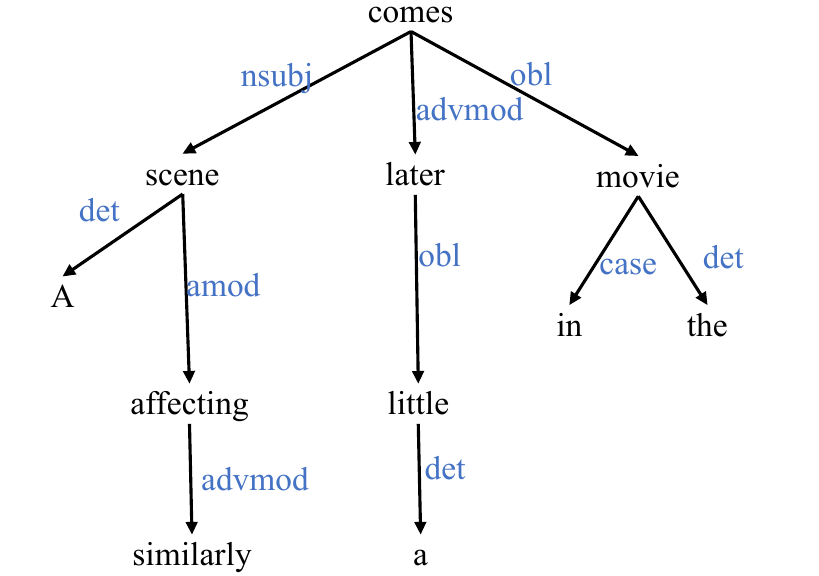}
	}
	
	\caption{Syntactic tree representation}
	\label{fig:syntactic-tree}
\end{figure}

A syntactic tree is a tree representation of different syntactic categories of a sentence and helps to understand the syntactic structure of a sentence.
There are two common syntactic tree structures: constituency syntactic and dependency syntactic\cite{manning2014stanford}, which are illustrated in Figure \ref{fig:syntactic-tree}.
Given a source sentence, a constituency syntactic tree presents a set of constituency relations, which shows how a word or group of words form different units within a sentence,
while a dependency syntactic tree presents a set of relations describing the direct relationships between words rather than how words constitute a sentence.
Dependency syntactic trees represent the grammatical relations that hold between constituents. 
Compared to constituency syntactic trees, dependency syntactic trees are more abstract, as they do not restrict or prescribe a particular word order. 
They are more specific in terms of semantics, and the notion of relations across words is explicit.

A dependency tree for a sentence is a directed acyclic graph with words as nodes and relations as edges. 
Each word in the sentence either modifies another word or is modified by a word. 
The root of the tree is the only entry that is modified but does not modify anything else. 
The relation that the two words participate in is given as a name on the edge connecting the nodes. 
More formally, the dependency structure tree can be expressed as follows: given a sentence $S=\{w_0, ..., w_n\}$, a set of edges or dependencies $E=\{e_1, ..., en\}$ are defined such that each $e_i$ connects two words in the sentence, and $w_0$ is a root word that only connects a word to another word.

\begin{figure*}[t]
\centering
\graphicspath{{graphs/}}
    \includegraphics[width=0.99\linewidth]{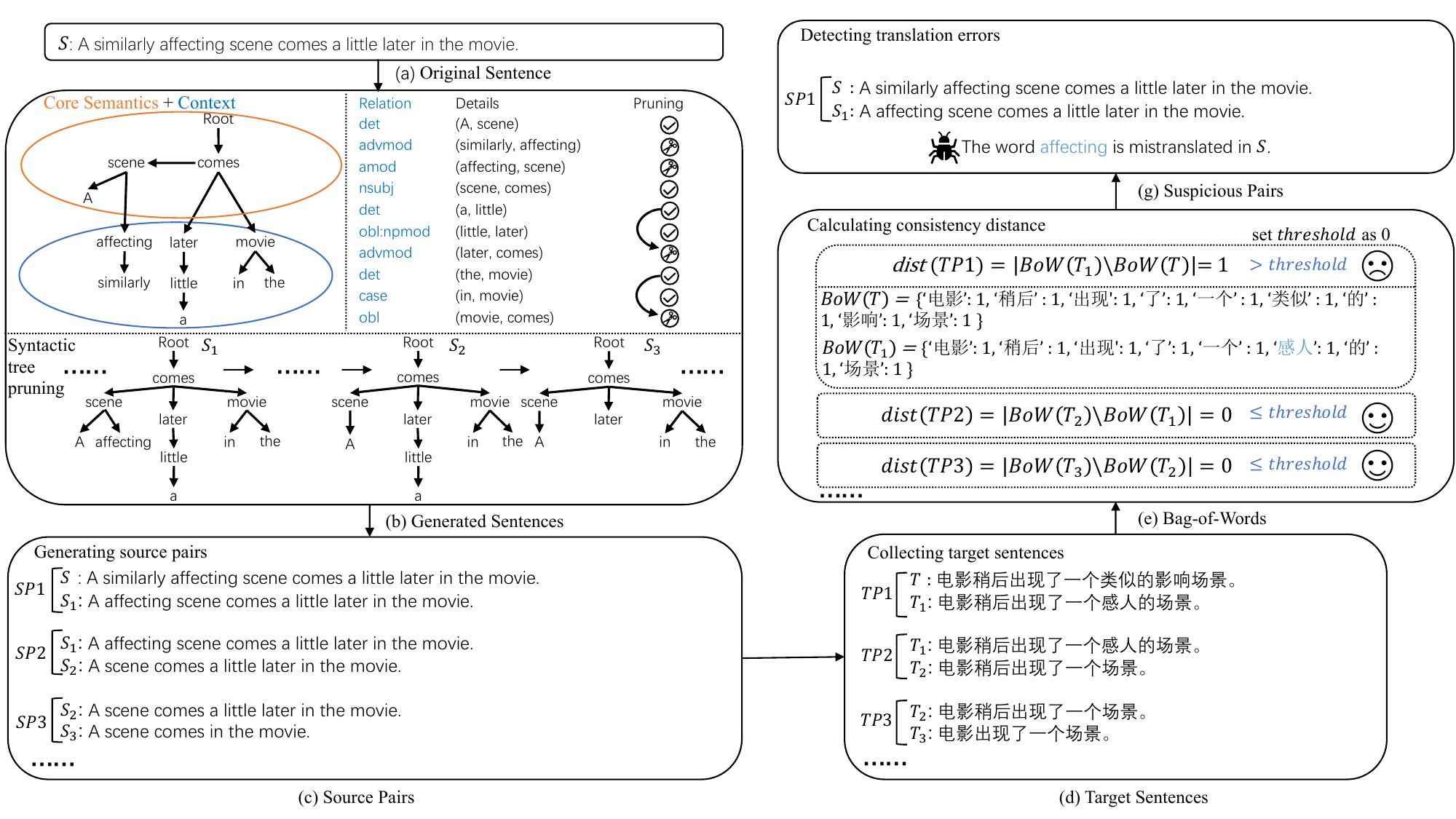}
    \caption{Overview of {\toolname}}
    \label{fig:approach}
\end{figure*}

\subsection{A Motivating Example}
\label{section: a_motivating_example}

In this section, we will present a real-world
erroneous translation example shown in Figure \ref{fig:approach} and illustrate how it is detected by syntactic tree pruning.
As a non-native English speaker, Echo often adopts machine translation systems to read news from other countries.
Echo reads a review article about the movie ``Single All the Way'' from CNN news website\footnote{https://edition.cnn.com/2021/12/02/entertainment/single-all-the-way-race-deconstructed-newsletter/index.html. Accessed August, 2022.}, and sees the the following English sentence (i.e., $S$ in Figure \ref{fig:approach}):

\begin{center}
\fbox{%
  \parbox{9cm}{%
    \begin{center}
      A similarly affecting scene comes a little later in the movie. 
    \end{center}
  }%
}
\end{center}

To figure out its meaning, Echo uses Google Translate, a popular translation service powered by neural machine translation models. 
Google Translate returns a corresponding Chinese sentence (i.e., $T$ in Figure \ref{fig:approach}).
\begin{center}
\fbox{%
  \parbox{9cm}{%
    \begin{center}
      \begin{CJK}{UTF8}{gbsn} 电影稍后出现了一个类似的影响场景。\end{CJK} 
    \end{center}
  }%
}
\end{center}

However, Echo finds he misunderstands the article because of the incorrect translation returned by Google Translate.
That is a real-world erroneous translation that leads to a confusing, unpleasant reading experience.
The correct translation should be:

\begin{center}
\fbox{%
  \parbox{9cm}{%
    \begin{center}
      电影稍后出现了一个类似的{\textcolor{red}{感人}}场景。 
    \end{center}
  }%
}
\end{center}

It is difficult for most existing \delete{technqiues }\revise{techniques} to find the translation error.
For example, the most recent technique CAT generates new sentences by replacing some words with other semantically similar words (e.g., $S'$: ``comes'' $\rightarrow$ ``occurs'', $S''$: ``comes'' $\rightarrow$ ``appears'' and $S'''$: ``movie'' $\rightarrow$ ``film'').
However, CAT fails to detect the erroneous translation as the generated sentences have the same translation as the original sentence.
Similar to CAT, SIT also performs word replacement to form a sentence pair (e.g., ``scene'' $\rightarrow$ ``moment'').
The pair is considered to have an erroneous translation if a large difference exists between the structures of translations in the pair.
However, Google Translate returns the translations with the same \delete{sentences }\revise{sentence} structures and the erroneous translation is also beyond the scope of SIT.

In fact, more and more people are getting used to relying on machine translation systems in their daily lives.
The robustness of the machine translation system is particularly crucial in a practical scenario.
It is observed that machine translation systems often make mistakes for complex sentences and perform well for simple sentences \cite{2020Sikka}.
Inspired by this, we propose to extract the core \delete{semantic }\revise{semantics} of a sentence and leverage the pruned sentence to test machine translation systems.
For the example sentence mentioned above, {\toolname} first performs syntactic tree scanning to identify "A scene comes a little later" as the core semantics and the remaining words as the context (shown in the upper left part of Figure \ref{fig:approach}(b)).
Then {\toolname} generates new sentences by removing context words in turn, which are then used to compare with the original sentence (shown in the upper right part of Figure \ref{fig:approach}(b)).
For example, in Figure \ref{fig:approach}(c), $S_1$ can be generated by removing the adjunct "similarly", and the translation should not influence the trunk of the original sentence.
We observe their translations $T$ and $T_1$ have different meanings, which means the sentence pair $<S, S_1>$ has erroneous translations.
In fact, the word ``affecting'' is mistranslated in the original sentence $S$ and some generated \delete{sentence }\revise{sentences} $S'$, $S''$ and $S'''$ with the same sentence structure. 
However, the word is translated correctly if the adjunct ``similarly'' is removed.
Thus, in this work, we attempt to eliminate \delete{irrelevant }\revise{contextual} information from the original source sentence without influencing the core semantics meaning (i.e., basic sentence structure) and propose {\toolname}, a novel syntactic tree pruning methodology for testing machine translation systems with arbitrary source sentences.

\section{Approach and Implementation}
\label{sec:approach}
This section introduces syntactic tree pruning testing ({\toolname}) and describes our practical implementation. 
In general, {\toolname} takes an unlabeled original source sentence as input \delete{for each time }\revise{at a time} and outputs a list of suspicious issues.
Figure \ref{fig:approach} presents the overview of {\toolname} implementation.
{\toolname} carries out the following steps:
\begin{enumerate}
  \item \textit{Core semantics-preserving pruned sentences generation}.
  For each unlabelled original source sentence, we generate a list of new source sentences by pruning the context at syntactic tree representation (shown in Figure \ref{fig:approach}(b)).

  \item \textit{Metamorphism-based source pair generation}.
  We pair each source sentence with all the new sentences generated from it to form source sentence pairs (shown in Figure \ref{fig:approach}(c)).

  \item \textit{Consistency-based translation error detection}.
  The core semantics of the translated generated sentences is compared to the core semantics of the translated original sentence via a bag-of-words model (shown in Figure \ref{fig:approach}(d) and Figure \ref{fig:approach}(d)(e)).
  If there is a large difference between the core semantics, {\toolname} reports a potential translation issue (shown in Figure \ref{fig:approach}(g)).
  
\end{enumerate}

Algorithm \ref{alg:pruning} presents the process of syntactic tree pruning.
The algorithm takes a list of arbitrary sentences as input and returns a list of suspicious pairs as output.
The main implementation process of this algorithm is first to establish a syntactic tree representation (line 3-4) for each original sentence by a recent neural network-based parser implemented in Stanford CoreNLP library \cite{2021-CoreNLP} and then generate a list of new sentences \delete{by }\revise{via} a novel core semantics-preserving pruning strategy (lines 22-43).
Then the source sentence pairs are fed to the machine translation systems based on the metamorphic relation (lines 7 -14), which are used to report suspicious translation errors via consistency detection (lines 17-21). 

\subsection{Core Semantics-preserving Pruned Sentences Generation}

\label{generate-sentences}

Given an arbitrary sentence in the source language, {\toolname} aims to generate new syntactically and semantically valid sentences by removing words or phrases from the original sentence. 
However, it is not trivial to choose words or phrases from a sentence without missing any crucial semantics and even undermining the sentence validity.
A naive implementation may be done by recursively removing a word or phrase until a predefined threshold (e.g., the minimum length of the generated sentence).
Although this implementation surely can generate all possible valid pruned sentences, it leads to too many candidate sentences with a low acceptability rate.
For example, the sentence in Section \ref{section: a_motivating_example} has nine words and we can generate 604,800 ($10*9*8*7*6*5*4$) new sentences if the predefined threshold is set to 3, as each word removal leads to a new sentence.
In the real world, the sentences fed to the machine translation systems tend to be more complex (e.g., more than 19 words on average for the dataset \cite{2020He}), and this implementation is unaffordable in practice (discussed in Section \ref{sec:random_deletion}).

Determining which word to remove from the original sentence can be aided by understanding the semantics of the word.
Inspired by the advance in natural language processing (NLP), several semantics-based tasks (e.g., sentence simplification or compression) may be well suited for this task \cite{2020Sikka, 2020Li}.
Specifically, given a source sentence, we extract semantics relations between words or phrases on its syntactic tree structure.
As a part of the grammar, syntax refers to the set of rules in a natural language that governs the structure of a sentence, determining how each component, such as words, phrases, and clauses, forms into their super-ordinate components, until the formation of the sentence \cite{chomsky2002syntactic}.
When removing a word from the sentence, the effect of the removal operation on other words should be considered.
Thus, the dependency syntactic tree is employed in this paper, as it effectively reflects the changes in how words interact \cite{2018Zhangb}. 

However, after the given sentence is represented as a dependency syntactic tree, several challenges remain in the pruning process.
Now we discuss several problems of generating new sentences by removing irrelevant content while preserving core semantics on the syntactic tree level.
Firstly, it is not trivial to define the core semantics and \delete{irrelevant }\revise{its} contextual information for a source sentence.
Natural languages often have complex syntactic structures, and it is essential to consider which dependencies can be used as the key semantics of a sentence.
Mann et al. \cite{1988Mann} describe a rhetorical structure theory (RST), which specifies each sentence component as either a nucleus or a satellite.
The nucleus component embodies the central piece of information, whereas the role of the satellite is to specify the nucleus further.
In linguistics, most sentences are combined by basic clause patterns (e.g., subject-predicate-object) and adjuncts (e.g., the attributive, adverbial).
The former as nucleus can express the sentence meaning clearly and the latter as the satellite is used to express the meaning more accurately. 
Then, we define the core semantics and context as follows.
\begin{definition}
\textbf{Core \delete{semantic }\revise{semantics} and context:~}
Given a sentence $S =\{S_e, S_c\}$ where $S_e$ is a set of words denoting the core \delete{semantic }\revise{semantics} and $S_c$ is the remaining words \delete{denotes }\revise{denoting} the associated context.
A $s_e \in S_e$ is a triple $(s, v, o) \in BT$, where $BT= \{SV, SVO, SVC, SVOO, SVOC\}$ represents the set of basic sentence structures.
Hence, $s \in S$ denotes a subject, $v \in V$ a verb and $o \in \{O, C, OO, OC\}$ \revise{denotes} a direct or indirect object, or complement.
\end{definition}

For example, in Figure \ref{fig:approach}(b), the nodes circled in orange represent the core semantics and the nodes circled in blue represent the context. 
In that way, the source sentence is reduced to the basic structure as the core sentence and augmented with some adjuncts that disclose associated contextual information.
On syntactic tree representation, the basic clause patterns are the trunk of the sentence, and the adjuncts are the branches of the sentence.

Secondly, the number of words that can be removed is difficult to determine at each step, as some words are dependent, such as ``a little later'' in Figure \ref{fig:approach}. 
Although we could adopt a strategy of removing one word at each case recursively, this approach would result in a large number of syntactically invalid sentences. 
For example, we cannot only remove the word ``later'' from the example sentence ``A similarly affecting scene comes a little later in the movie.''.
When removing such words, we have three possible situations:
(1) Unprunable. 
The word constitutes the five basic clause types. 
The deletion of these words will result in the absence of core semantics.
(2) Prunable.
The word does not constitute the five basic clause types and the removal operation of the word does not affect the \newdelete{legality}\newrevise{validity} of the sentence.
(3) Partially-Prunable.
The word does not constitute the five basic clause types and only the removal operation of the word and other words it depends on does not affect the \newdelete{legality}\newrevise{validity} of the sentence.

\begin{algorithm}[t]
\caption{Implementation of {\toolname}}
\footnotesize
\label{alg:pruning}
\begin{algorithmic}[1]
    \renewcommand{\algorithmicrequire}{\textbf{Input:}}
    \renewcommand{\algorithmicensure}{\textbf{Output:}}
    \Require $source\_{sents}$: a list of sentences in the source language
    \Ensure  $suspicious\_{issues}$: a list of suspicious pairs 
    
    \State $suspicious\_{issues}  \gets  List()$
    \For{each $source\_{sent} \in source\_{sents}$}
        \State $dependency\_{tree} \gets PARSE(source\_{sent})$
        \State $head \gets dependency\_{tree}.head()$
        \State $gen\_{sents} \gets PRUNING(source\_{sent}, dependency\_{tree},head)$
        \State $target\_{source}\_{sent} \gets TRANSLATE(source\_{sent})$
        
      \For{each $gen\_{sent} \in gen\_{sents}$}
        \State $target\_{gen}\_{sent} \gets TRANSLATE(gen\_{sent})$
        
        \If{DISTANCE($target\_{source}\_{sent}, target\_{gen}\_{sent}) > d$}
            \State $source\_{pair} \gets \{source\_{sent}, gen\_{sent}\}$
            \State $target\_{pair} \gets \{target\_{source}\_{sent}, target\_{gen}\_{sent}\}$
        
            \State $suspicious\_{issues}.add(source\_{pair}, target\_{pair})$
        \EndIf
      \EndFor
    \EndFor
    \State \Return $suspicious\_{issues}$
     
    \
    
    \Function{DISTANCE}{$target\_{source}\_{sent}, target\_{gen}\_{sent}$}
    \State $source\_{BOW} \gets BAGOFWORDS(target\_{source}\_{sent})$
    \State $gen\_{BOW} \gets BAGOFWORDS(target\_{gen}\_{sent})$
    \State \Return $|source\_{BOW} \setminus gen\_{BOW}|$
    
    \EndFunction
    
    \
    
    \Function{PRUNING}{$source\_{sent},dependency\_{tree},head$}
    \If{$len(head.children()) > 0$}
        \State $nodes \gets head.children()$
        \State $dep\_nodes \gets DEP\_PRIOR(dependency\_{tree})$
        \Comment{prioritize the nodes in the dependency tree}
        \State $relation\_set \gets RELATION()$ 
        \Comment{predefined operation set}
        \For{each $node \in dep\_nodes$}
            \State $source\_sent \gets REMOVE($source\_sent, relation\_set, node$)$
            
            \If{$len(node.children()) > 0$}
            \State PRUNING($source\_sent,dependency\_tree,node$)
            \EndIf
            \If{node.is\_removed()}
            \State $gen\_sents.add(source\_sent)$
            \EndIf
        \EndFor
    \Else
    \State $source\_sent \gets REMOVE($source\_sent, relation\_set, head$)$
    \Comment{determine whether to delete the node according to the relation dictionary}
    \If{head.is\_removed()}
    \State $gen\_sents.add(source\_sent)$
    \EndIf
    \EndIf
    
    \State \Return $gen\_{sents}$
    \EndFunction
    
\end{algorithmic}
\end{algorithm}

To address the above challenges,
 we perform a four-step cascaded pruning strategy: sentence type identification, sentence simplification, core semantics extraction and pruned sentence generation.
(1) {\toolname} performs a syntactic tree scanning and divides the given original sentence into four types, i.e., simple, compound, complex and compound-complex sentences based on the edges connecting the nodes.
The type of sentence is determined by how many clauses, or subject-verb groups, are included in the sentence.
(2) After determining the sentence type, {\toolname} attempts to split the compound and complex sentence into simple sentences.
For example, a compound sentence will be split according to the co-dependency relation (\newrevise{i.e.,} \textit{cc} edge).
The sub-trees before and after the edge are divided into new sentences at the same time, and the process is recursive until no compound sentence exists.
The noun clause in complex sentences will also be split according to the dependency relation, and the new sentence after splitting only shares conjunctions. 
The remaining clauses in complex sentences will not affect the core semantics of the sentence, so they will be pruned.
(3) For each simple sentence, {\toolname} identifies the shortest path among the syntactic tree, which is used to represent the crucial semantics of the original sentence.
As most sentences are derived from the five basic sentence structures (i.e., subject–verb, subject–verb–object, subject–verb–complement, subject–verb–object-object, subject–verb–object-complement), the nodes constituting the structures are considered as the shortest path.
\newrevise{Thus, although the generated sentence has a different structure from the original one, they share the same basic structure.}
(4) {\toolname} performs a removal operation for each node according to the edges (i.e., dependencies) until the shortest path remains.
Among the nodes, there are some constraints between them.
In the syntactic tree, when pruning a specific node, its dependent node also needs to be considered.
We consider all available dependency types implemented in Stanford’s CoreNLP library\footnote{English dependencies description. https://downloads.cs.stanford.edu/nlp/software/dependencies\_manual.pdf. Accessed August, 2022.} and manually analyze the dependency structure trees on 1000 sentences randomly selected from the news \cite{2020He}.
We then determine which of the above possible situations each dependency type belongs to and design the corresponding pruning operator.
Specifically, the unprunable situation maps $up$ operator, which denotes the node should be retained, as it belongs to the five basic clause types.
The prunable situation maps $pr$ operator, which denotes the node can be deleted, as it does not belongs to the five basic clause types and not affect the \newdelete{legality}\newrevise{validity} of the sentence.
The partially-prunable situation maps $pp$, which denotes the node should be deleted with its dependent node simultaneously, as only deleting both of them does not affect the \newdelete{legality}\newrevise{validity} of the sentence.
With these corresponding operators, we establish a mapping table to determine whether the dependent node needs to be cascaded.
Details of the dependencies and the corresponding map table are presented in Table \ref{TAB:mapping}.
Table \ref{TAB:mapping} contains four main columns, each of which has two sub-columns.
The first sub-column lists the dependency types.
The second sub-column lists the three types of pruning operators (i.e., $up$, $pr$ and $pp$).
Through the mapping table, we can determine the least nodes that can be pruned each time.
For example, given a simple sentence $S$ in Figure \ref{fig:approach},  we first identify the crucial semantics (i.e., the circled part) and then iterate over all nodes to determine if they should be deleted based on Table \ref{TAB:mapping}.

\begin{table}[htbp]
    \footnotesize 
\caption{Dependency relation mapping table.}
\label{TAB:mapping}
 \begin{tabular}{ll|ll|ll|ll}
\toprule
Relation & Type & Relation & Type & Relation & Type & Relation & Type\\
\midrule
    ROOT  & $up$     & iobj  & $up$     & predet & $pr$     & poss  & $pr$ \\
    dep   & $pp$     & nsubj & $up$     & preconj & $pr$     & prt   & $pp$ \\
    aux   & $pp$     & dobj  & $up$     & mwe   & $pp$     & compound & $pr$ \\
    auxpass & $pp$     & det   & $pp$     & mark  & $pp$     & goeswith & $pp$ \\
    cop   & $up$     & expl  & $pp$     & advmod & $pr$     & ref   & $pp$ \\
    ccomp & $pp$     & amod  & $pr$     & neg   & $pr$     & xsubj & $up$ \\
    xcomp & $pp$     & nmod  & $pr$     & tmod  & $pr$     & case  & $pp$ \\
    obj   & $up$     & nummod & $pp$     & punct & $up$     & obl & $pr$ \\
\bottomrule
\end{tabular}
\end{table}

\subsection{Metamorphism-based Source Pair Generation}
\label{sec:generating-pairs}

Once all possible pruned sentences have been generated, they should be translated into the target language for semantics validation (Section \ref{detect-errors}).
{\toolname} is designed to perform the metamorphic testing for the machine translation systems $M$.
The most significant advantage of metamorphic testing lies in its capability of identifying oracle information for the tests via metamorphic relations (MR).
Given a source sentence set S and the pruning operators $P$, the MR \delete{can be }\revise{is} defined as a set of sentences that can be derived from each sentence $s \in S$ without influencing the crucial information. 
This MR to test $M$ with additional transformed sentences can be formalized as follows:

\begin{equation}
\forall s \in S \wedge \forall p \in P, \quad M(s)=M(p(s))
\end{equation}

Given this MR, we can simply obtain the oracle information by verifying whether it is satisfied in the testing process of $M$.
Under this setting, each generated sentence must be paired with the original sentence, which will be used to check whether the relation is satisfied or violated.
If a violation is detected, we can then say that $M$ is faulty.
Specifically, each source sentence pair should have two different pieces of text that contain the same phrase.
To generate these pairs, we pair each source sentence (i.e., original and generated) with all the generated sentences pruned from the source sentence.
For example, as illustrated in Figure \ref{fig:approach}, source sentence $S1$ is pruned from the original source sentence $S$ by removing the word ``similarly''.
Meanwhile, new source sentence $S2$ can also be pruned from the generated source sentence $S1$ by removing the phrase ``affecting''.
Thus, three source sentence pairs will be constructed: (1) the original sentence $S$ and the generated source sentence $S1$; (2) the generated sentence $S1$ and the generated source sentence $S2$; and (3) the generated sentence $S2$ and the generated source sentence $S3$.

\subsection{Consistency-based Translation Error Detection}
\label{detect-errors}

After pruned source sentences have been generated, we feed them and the corresponding original source sentences to the machine translation systems under test.
For each original source sentence and all generated source sentences, each machine translation system takes a source sentence and a target language (i.e., Chinese) as \delete{input }\revise{inputs}, and then returns a translation sentence in a target language.

Once the target sentences are collected, we need to detect erroneous translations.
It is non-trivial to detect whether a pruned sentence preserves the core semantics of the original sentence without reporting false positives.
Given an original source sentence $E=\{e_1, e_2, \cdots, e_n\}$, we generate a corresponding pruned source sentence $E'=\{e'_1, e'_2, \cdots, e'_m\}$ by removing several words, and their translations are $T(E)$ and $T(E')$.
It is challenging to map the relation between the source language pair (i.e., $E$ and $E'$) to the target language pair (i.e., $T(E)$ and $T(E')$).
Machine translation systems may return a sentence with a different structure for imperceptible perturbations due to the brittleness of the neural network.
Thus, following existing work \cite{2021He}, we adopt a bag-of-words (BoW) model, a simple representation disregarding grammar and even word order but keeping multiplicity, to capture the inconsistent semantics between the original and pruned sentences.
Although simple, it has been proven quite effective in some NLP tasks, such as neural machine translation \cite{2010Wu, 2021He}.
For each translated pairs $T(E)$ and $T(E')$,
the distance between $T(E)$ and $T(E')$ is calculated by 

\begin{equation}
    dist(T(E), T(E')) = |BoW(T(E')) \setminus BoW(T(E))|	
\end{equation}

where $BoW(T(E))$ and $BoW(T(E'))$ denote the BoW representation of $T(E)$  and $T(E')$.
The $\setminus$ operator denotes the set difference (i.e., how many word occurrences in $BoW(T(E'))$ but not in $BoW(T(E)$.
For example, the distance between ``A similarly affecting scene comes a little later in the movie'' and  ``A affecting scene comes a little later in the movie'' is 1.
\revise{
Following most existing machine translation testing work \cite{2020He, 2020Sun, 2019WangWenyu, 2019Zheng, 2020Gupta, 2021He, sun2022improving}, we report suspicious issues via a pre-defined threshold value $t$.
}
If the distance is larger than \delete{a }\revise{the} threshold \revise{value} $t$, the translated pair is considered to break the consistency property and will be reported as a suspicious issue.
For example, $t=0$ represents the distance of all reported issues is larger than \revise{the} threshold $0$, i.e., there exist \delete{words }\revise{new word occurrences} in \revise{the} pruned translation but not in \revise{the} original translation.
\revise{
The threshold value demonstrates the degree of inconsistency in core semantics between the original and pruned sentences, and controls the trade-off between the precision and the number of reported suspicious issues.
When we increase the threshold value $t$, there exist more new words introduced in the translation of the pruned sentence, leading to more accurate reported issues.
Such a threshold setting is also adopted in previous work\cite{2020He, 2020Gupta} and proves its effectiveness in detecting translation errors.
}
Specifically, for each original source sentence, {\toolname} reports either no issue or a list of suspicious issues.
There may exist three possible translation error types in an issue: 
(1) the original source sentence has an erroneous translation, (2) the generated source sentence has an erroneous translation, and (3) both the source sentences have erroneous translations.
A suspicious issue consists of the original source sentence, a generated source sentence, and their translations.

\section{Experimental Setup}
\label{sec:exp}
\subsection{Research Questions}
Our main research questions are as follows.
\begin{description}
    \item[RQ1.] How precise is {\toolname} at finding erroneous issues?

    \item[RQ2.] How many erroneous translations \delete{can }\revise{does} {\toolname} report?
    
    \item[RQ3.] What type of erroneous translations \delete{can }\revise{does} {\toolname} report?
    
    \item[RQ4.] How efficient is {\toolname} in terms of running time?
    
\end{description}

\subsection{Machine Translation Systems}

Our experiment considers two state-of-the-art industrial machine translators.
The former is Google Translate, and the latter is Bing Microsoft Translator \cite{Google, Microsoft}.
Both of them are widely used machine translation services developed by Google and Microsoft, and have been widely adopted in recent machine translation testing work \cite{2020Gupta,2020He,2020Sun,2021He}.
Specifically, we invoke the APIs provided by Google Translate and Bing Microsoft Translator to obtain the translation results,
which are identical to those returned by their Web interfaces.

\subsection{Dataset}
\label{sec:dataset}

\begin{table}[t]
    \footnotesize 
    \centering
    \caption{Statistics of input sentences for evaluation}
    \label{TAB:Dataset}
    \begin{tabular}{ccccc}
    \toprule
    Corpus & \# of words/sentence & Average \# of words/sentence & Total \# of words & Distinct \# of words\\
    \midrule
    Politics & 12-69 & 23.82 & 2,382 & 1,089 \\
    Business & 11-49 & 25.85 & 2,585 & 1,109 \\
    Culture & 11-49 & 24.02 & 2,402 & 1,160 \\
    Sports & 11-49 & 23.83 & 2,383 & 1,035 \\
    Tech & 11-38 & 21.17 & 2,117 & 1,009 \\
    Travel & 10-63 & 21.74 & 2,174 & 1,018 \\
    Health & 11-47 & 22.30 & 2,230 & 877 \\
    Life & 11-46 & 24.88 & 2,488 & 1,112 \\
    Legal & 11-52 & 23.74 & 2,374 & 967 \\
    Opinion & 11-48 & 22.07 & 2,207 & 796 \\
    \midrule
    Politics* & 4-32 & 19.20 & 1,918 & 933 \\
    Business* & 4-33 & 19.50 & 1,949 & 944 \\
    \bottomrule
    \end{tabular}
\end{table}

Following previous machine translation testing work \cite{2020He, 2020Gupta}, we collect real-world source sentences from the news.
Specifically, we randomly collect \revise{the} latest articles from CNN (Cable News Network) \cite{2021cnn}, China Daily \cite{2021chinadaily}, BBC (British Broadcasting Corporation) \cite{2021bbc} and Reuters \cite{2021reuters}.
To evaluate whether {\toolname} consistently performs well on sentences of different semantic contexts,
the \delete{article }\revise{articles} are extracted from ten categories: Business, Culture, Lifestyle, Sports, Politics, Travel, Technology, Healthy, Opinion and Legal.
For each category, we extract its main text contents, and split them into a list of sentences.
Then we randomly select 100 syntactically and semantically correct sentences from each category.
We also experiment on \delete{the }\revise{a widely-adopted} dataset from previous machine translation testing work \cite{2020He, 2020Gupta, 2021He} containing 200 sentences.
Specifically, the dataset consists of articles from two categories: Politics and Business, where each dataset contains 100 English sentences.
In total, we have 1,200
sentences from ten major categories of news sites \revise{as the initial sentence corpus for {\toolname} and other baselines} in the experiment.
To the best of our knowledge, this is the largest evaluation dataset for machine translation testing so far.

Statistics of the dataset \delete{is }\revise{are} illustrated in Table \ref{TAB:Dataset}.
The first column lists the twelve dataset categories, where the first ten rows denote the sentence collected in our work and the last two rows denote the sentences from He et al. \cite{2020He, 2020Gupta, 2021He}.
The second column list\revise{s} the minimum-maximum number of words and the third column list\revise{s} the average number of words for all sentences in the category.
Similarly, the fourth column lists the total number of words that appear in all sentences.
The fifth column lists the number of non-repetitive words in all sentences.
For example, sentences in Culture dataset contain 11-49 words (the average is 24.02 words) and they contain 2,402 words and 1,160 non-repetitive words in total.

\subsection{Labelling}
\label{sec:labelling}
An issue contains a pair of source sentences, where the first one is the original sentence and the second is the generated sentence.
To construct the \delete{label }\revise{labelling} process, we invite 20 graduate students in our experiment.
All of them have more than 10 years of English learning experience and are all native speakers of Chinese.
\revise{
Before the formal labelling process, we provide documents and presentation videos as a guide to all participants, to ensure they fully understand the labelling procedure.}
Specifically, the participants identify whether the source sentences in the reported issue contain any syntax or semantic error.
If both source sentences are invalid in the issue, it is labeled as a false positive.
Then, for the remaining issues, the participants check whether there is an erroneous translation for the valid source sentence(s).
If so, the participants count the reported issue as a true positive.
Otherwise, the reported issue is labeled as a false positive.
Finally, the participants decide which source sentence(s) in this issue cause(s) the translation error(s).

As we manually label the issues, it is inevitable to introduce subjectivity.
To minimize such subjectivity, we \delete{assigned }\revise{assign} each issue to two different participants. 
When there \delete{was }\revise{exists} a disagreement, all the participants would involve to have an open discussion to resolve it\footnote{The Cohen'Kappa score is 0.93 on average.}. 
Moreover, we \delete{mixed }\revise{mix} issues reported by our system and the ones from other systems, and thus the participants \delete{were }\revise{are} unaware of whether an issue \delete{was }\revise{is} reported using our system or not.
\revise{Thus, we believe that the human labelling process can provide reliable ground truth information in our experiment.}

\subsection{Comparison}
\label{sec:baseline}
To evaluate the performance of {\toolname}, following the most recent work \cite{sun2022improving}, we compare it with five state-of-the-art machine translation testing techniques: \textbf{SIT}, \textbf{TransRepair}, \textbf{RTI}, \textbf{PatInv} and \textbf{CAT}.
To our knowledge, our work is the largest evaluation study on machine translation testing so far.
We obtain the source code of the first four techniques from the TestTranslation toolkit \cite{toolkit} and CAT from the previous study \cite{sun2022improving}.

Specifically, TransRepair adopts four similarity metrics (i.e., LCS-based metric, ED-based metric, tf-idf-based metric and BLEU-based metric) for measuring inconsistency.
Following existing work \cite{2021He}, we select ED-based metric (referred \revise{to} as TransRepair-ED) in our study because it achieves the highest precision and detects the second largest number of translation errors among four metrics on Google Translate.
PatInv has two variants to generate sentences with \revise{a} different meaning, i.e., replacing one word in a sentence with a non-synonymous word (referred \revise{to} as PatInv-Replace) and removing a meaningful word or phrase from the sentence (referred \revise{to} as PatInv-Remove).
In this work, we select PatInv-Replace because it performs significantly better than PatInv-Remove in terms of precision and the number of erroneous translations.
SIT adopts three different metrics (i.e., edit distance, constituency set distance, and dependency set distance) for evaluating the distance between target sentences.
We select \revise{the} dependency set distance metric (referred \revise{to} as SIT-Dep) in our study as it has the best performance on top-1 precision for both Google Translate and Bing Microsoft Translator \cite{2020He}.
CAT identifies word replacement with isotopic replacement and adopts the same similarity metrics in TransRepair.
Following the original work \cite{sun2022improving}, we use LCS-based metric to represent CAT (referred \revise{to} as CAT-LCS) in our evaluation.

\subsection{Experimental Environments}
\label{sec:hardware}

All experiments \revise{(including {\toolname} and baselines)} are conducted on Ubuntu 16.04 with 16GB RAM and 6 Intel Core i7-10710U CPUs.
For sentence parsing, we adopt a shift-reduce parser \cite{2013Zhu}
which is implemented in Stanford CoreNLP toolkit \cite{2021-CoreNLP}.
Our experiments consider the English $\rightarrow$ Chinese language setting because of the knowledge background of the authors.

\section{Evaluation}
\label{sec:eva}

\setlength{\rotFPtop}{0pt plus 1fil}
\setlength{\rotFPbot}{0pt plus 1fil}
\hspace*{-6pt}
\begin{sidewaystable}[!tp]
    \caption{The precision of {\toolname} (\# of erroneous issues/\# of suspicious issues) under different threshold values}
    \label{tab:precision}
    \scriptsize
    \setlength\tabcolsep{0.75pt}
    \begin{tabular}{l|ccccccc|c|c|c|c|c}
    \toprule
         & \multicolumn{7}{c|}{{\toolname}} & \multirow{2}{*}{SIT} & \multirow{2}{*}{PatInv} & \multirow{2}{*}{RTI} & \multirow{2}{*}{TransRepair} & \multirow{2}{*}{CAT}\\
         \cline{2-8}
         & $t=0$ & $t=2$ & $t=4$ & $t=6$ & $t=8$ & $t=10$ & $t=12$ & & & & &\\
    \midrule
    Google Politics & 61.0\%(1,410/2,310) & 68.8\%(645/937) & 76.7\%(207/270) & \textbf{71.4\%(35/49)} & 77.8\%(7/9) & N.A.  & N.A.  & 53.8\%(7/13) & 64.3\%(36/56) & 37.1\%(49/132) & 64.7\%(11/17) & 62.3\%(134/215) \\
    Google Business & 86.1\%(2,895/3,364) & 85.1\%(1,238/1,455) & 79.3\%(298/376) & \textbf{76.3\%(58/76)} & 90.0\%(18/20) & 100.0\%(8/8) & 100.0\%(6/6) & 61.1\%(11/18) & 35.6\%(21/59) & 39.0\%(55/141) & 31.6\%(12/38) & 52.2\%(109/209) \\
    Google Culture & 60.5\%(2,574/4,255) & 59.6\%(1,357/2,278) & 61.5\%(606/985) & \textbf{62.2\%(153/246)} & 67.5\%(27/40) & 80.0\%(4/5) & N.A.  & 29.4\%(5/17) & 34.0\%(17/50) & 32.6\%(61/187) & 56.0\%(14/25) & 37.9\%(89/235) \\
    Google Sport & 55.7\%(1,315/2,361) & 58.6\%(675/1,151) & 60.7\%(244/402) & \textbf{80.1\%(125/156)} & 93.4\%(71/76) & 100.0\%(38/38) & 100.0\%(21/21) & 44.0\%(11/25) & 73.7\%(216/293) & 48.3\%(57/118) & 70.2\%(33/47) & 35.3\%(84/238) \\
    Google Tech & 63.1\%(1,283/2,033) & 74.9\%(590/788) & 78.0\%(206/264) & 71.0\%(44/62) & 92.3\%(12/13) & N.A.  & N.A.  & 28.6\%(4/14) & 55.6\%(15/27) & 26.9\%(21/78) & 40.0\%(6/15) & \textbf{87.5\%(182/208)} \\
    Google Travel & 67.2\%(3,087/4,591) & 73.2\%(1,816/2,480) & 80.1\%(859/1,073) & \textbf{86.6\%(335/387)} & 90.6\%(125/138) & 90.6\%(48/53) & 95.5\%(21/22) & 22.2\%(2/9) & 68.8\%(11/16) & 30.9\%(42/136) & 55.6\%(10/18) & 69.7\%(182/261) \\
    Google Health & 51.0\%(1,579/3,099) & 60.8\%(921/1,514) & 69.8\%(388/556) & \textbf{84.9\%(90/106)} & 100.0\%(24/24) & 100.0\%(7/7) & N.A.  & 71.4\%(10/14) & 37.5\%(15/40) & 50.0\%(29/58) & 63.2\%(12/19) & 72.4\%(152/210) \\
    Google Life & 55.8\%(1,837/3,295) & 62.9\%(1,098/1,745) & 80.1\%(447/558) & \textbf{92.5\%(149/161)} & 100.0\%(26/26) & 100.0\%(3/3) & N.A.  & 27.8\%(5/18) & 46.2\%(18/39) & 37.0\%(44/119) & 35.5\%(11/31) & 63.7\%(142/223) \\
    Google Legal & 62.2\%(1,848/2,971) & 73.0\%(1,069/1,464) & 79.0\%(459/581) & \textbf{83.3\%(169/203)} & 72.4\%(42/58) & 54.5\%(6/11) & 66.7\%(2/3) & 58.8\%(10/17) & 29.2\%(14/48) & 40.2\%(39/97) & 33.3\%(6/18) & 72.8\%(139/191) \\
    Google Opinion & 61.1\%(1,553/2,541) & 70.0\%(700/1,000) & 77.5\%(172/222) & \textbf{85.0\%(51/60)} & 100.0\%(17/17) & 100.0\%(2/2) & N.A.  & 43.8\%(7/16) & 41.3\%(19/46) & 38.4\%(48/125) & 55.6\%(10/18) & 51.2\%(110/215) \\
    Google Politics* & 82.5\%(933/1,131) & 89.0\%(389/437) & 97.6\%(120/123) & \textbf{100.0\%(38/38)} & 100.0\%(12/12) & 100.0\%(2/2) & N.A.  & 82.7\%(43/52) & 71.3\%(77/108) & 74.5\%(73/98) & 77.6\%(83/107) & 58.4\%(125/214) \\
    Google Business* & 87.2\%(1,094/1,254) & 94.5\%(464/491) & 97.3\%(109/112) & \textbf{100.0\%(14/14)} & N.A.  & N.A.  & N.A.  & 74.5\%(38/51) & 79.8\%(75/94) & 70.7\%(58/82) & 79.6\%(74/93) & 75.0\%(168/224) \\
    \midrule
    \multirow{2}{*}{Google Sum} & 64.5\% & 69.6\%($\uparrow$ 5.1) & 74.5\%($\uparrow$ 10.0) & \textbf{80.9\%($\uparrow$ 16.4)} & 88.0\%($\uparrow$ 23.5) & 91.5\%($\uparrow$ 27.0) & 96.2\%($\uparrow$ 31.7) & 58.0\% & 61.0\% & 42.0\% & 63.2\% & 61.1\% \\
     & (2,1,444/3,3,222) & (1,0,999/1,5770) & (4,111/5,522) & (1,266/1,555) & (381/433) & (118/129) & (50/52) & (153/264) & (534/876) & (576/1,377) & (282/446) & (1,611/2,644) \\
    \midrule

    Bing Politics & 72.9\%(1,688/2,311) & 75.5\%(707/937) & 75.6\%(204/270) & \textbf{75.5\%(37/49)} & 55.6\%(5/9) & N.A.  & N.A.  & 61.5\%(8/13) & 41.4\%(12/29) & 40.0\%(52/130) & 68.8\%(11/16) & 62.6\%(129/206) \\
    Bing Business & 66.6\%(2,122/3,199) & 72.1\%(941/1,300) & 73.3\%(217/296) & \textbf{73.0\%(27/37)} & 100.0\%(5/5) & N.A.  & N.A.  & 61.1\%(11/18) & 12.0\%(6/50) & 36.2\%(59/163) & 44.0\%(11/25) & 54.6\%(112/205) \\
    Bing Culture & 75.1\%(3,122/4,166) & 82.8\%(1,755/2,122) & 84.4\%(777/921) & \textbf{79.9\%(270/338)} & 76.4\%(68/89) & 76.2\%(16/21) & 88.9\%(8/9) & 29.4\%(5/17) & 15.0\%(3/20) & 48.9\%(86/176) & 40.9\%(9/22) & 62.5\%(140/224) \\
    Bing Sport & 51.5\%(1,199/2,322) & 64.8\%(712/1,099) & 77.3\%(307/397) & \textbf{88.1\%(126/143)} & 95.2\%(59/62) & 95.8\%(23/24) & 100.0\%(6/6) & 34.6\%(9/26) & 25.6\%(23/90) & 55.3\%(73/132) & 70.0\%(21/30) & 74.5\%(146/196) \\
    Bing Tech & 64.8\%(1,333/2,066) & 79.9\%(609/762) & 90.1\%(192/213) & \textbf{100.0\%(49/49)} & 100.0\%(7/7) & N.A.  & N.A.  & 57.1\%(8/14) & 44.4\%(4/9) & 30.0\%(21/70) & 38.9\%(7/18) & 81.5\%(181/222) \\
    Bing Travel & 69.9\%(3,188/4,555) & 73.3\%(1,800/2,466) & 80.7\%(820/1,011) & \textbf{87.3\%(261/299)} & 91.1\%(72/79) & 100.0\%(9/9) & 100.0\%(2/2) & 44.4\%(4/9) & 56.2\%(9/16) & 25.4\%(29/114) & 47.1\%(8/17) & 72.7\%(157/216) \\
    Bing Health & 51.1\%(1,555/3,033) & 53.8\%(792/1,477) & 55.2\%(267/484) & 64.4\%(56/87) & 50.0\%(3/6) & N.A.  & N.A.  & \textbf{78.6\%(11/14)} & 32.4\%(11/34) & 59.7\%(37/62) & 64.7\%(11/17) & 70.6\%(125/177) \\
    Bing Life & 66.6\%(2,122/3,199) & 72.4\%(1,066/1,477) & 83.2\%(356/428) & \textbf{85.4\%(76/89)} & 76.9\%(10/13) & 100.0\%(1/1) & N.A.  & 50.0\%(9/18) & 25.9\%(7/27) & 37.0\%(47/127) & 50.0\%(17/34) & 55.0\%(116/211) \\
    Bing Legal & 60.7\%(1,755/2,888) & 70.0\%(861/1,233) & 84.9\%(382/450) & \textbf{92.8\%(193/208)} & 96.1\%(98/102) & 93.0\%(53/57) & 92.2\%(47/51) & 70.6\%(12/17) & 36.6\%(15/41) & 49.5\%(54/109) & 44.4\%(8/18) & 79.3\%(146/184) \\
    Bing Opinion & 60.1\%(1,533/2,555) & 60.3\%(551/914) & 61.7\%(153/248) & \textbf{73.5\%(25/34)} & 100.0\%(7/7) & 100.0\%(1/1) & 100.0\%(1/1) & 43.8\%(7/16) & 42.4\%(25/59) & 40.4\%(55/136) & 66.7\%(8/12) & 46.9\%(99/211) \\
    Bing Politics* & 80.7\%(837/1,033) & 86.9\%(337/388) & 95.8\%(92/96) & \textbf{100.0\%(17/17)} & 100.0\%(3/3) & N.A.  & N.A.  & 84.3\%(43/51) & 60.8\%(48/79) & 58.6\%(65/111) & 53.8\%(43/80) & 63.7\%(114/179) \\
    Bing Business* & 67.7\%(831/1,222) & 76.0\%(336/442) & 85.5\%(65/76) & \textbf{100.0\%(5/5)} & N.A.  & N.A.  & N.A.  & 75.0\%(33/44) & 42.2\%(43/102) & 68.4\%(78/114) & 67.3\%(68/101) & 70.9\%(139/196) \\
    \midrule
    \multirow{2}{*}{Bing Sum} & 65.4\% & 71.7\%($\uparrow$ 6.3) & 78.3\%($\uparrow$ 12.9) & \textbf{84.3\%($\uparrow$ 18.9)} & 88.2\%($\uparrow$ 22.8) & 91.2\%($\uparrow$ 25.8.1) & 92.8\%($\uparrow$ 27.4) & 62.3\% & 37.1\% & 45.4\% & 56.9\% & 66.1\% \\
     & (2,1229/3,2556) & (1,0448/1,4663) & (3,833/4,899) & (1,144/1,355) & (337/382) & (103/113) & (64/69) & (160/257) & (206/556) & (656/1,444) & (222/390) & (1,600/2,422) \\
    \bottomrule
    
    \multicolumn{10}{l}{\small 1. $\uparrow$ denotes performance improvement against STP with $t=0$.} \\
    \multicolumn{10}{l}{\small 2. The bold cell denotes optimal precision among STP with $t=6$, SIT, PatInv, RTI, TransRepair and CAT.}
    \end{tabular}
\end{sidewaystable}

\subsection{Precision on Finding Erroneous Issues}
\label{sec:precision}

{\toolname} aims to automatically detect erroneous issues for machine translation systems using arbitrary sentences.
Thus, we evaluate the effectiveness of {\toolname} from two aspects: 
(1) how many erroneous issue {\toolname} \delete{can }\revise{is able to} report;
and (2) the precision of the erroneous issue reported by {\toolname}.

\subsubsection{Evaluation Metric}
Given a list of unlabeled, monolingual source language sentences, the output of {\toolname} is a list of suspicious issues $\mathfrak{}{I}$.
A suspicious issue $\mathfrak{i}$ contains 
(1) an original sentence, $E$, in the source language, and its translation sentence, $T(E)$, in target language; 
(2) a pruned sentence, $E'$, in the source language, and its translation sentence, $T(E')$, in target language.
We define the precision as the percentage of erroneous issues, where there exist translation error(s) in $T(E)$ or $T(E')$.
Formally, for a suspicious issue $p$, we set $error(p)$ to $true$, if $T(E)$ or $T(E')$ has translation error(s), otherwise we set it to $false$.
Given a list of suspicious issues, the precision is calculated by the number of \delete{the erroneous problems }\revise{erroneous issues} divided by the total number of reported \revise{suspicious} issues:

\begin{equation}
\small
    Precision=\frac
    {\sum_{p \in P} \mathbbm{1} \{error(p)\}}
    {|P|}
\label{equation:precision}
\end{equation}
where $\mathbbm{1}$ denotes whether a suspicious issue $p$ contains translation error(s) and  $|P|$ denotes the number of suspicious issues returned by {\toolname}.

\subsubsection{Results}
The details of the comparison results are presented in Table \ref{tab:precision}.
The first column lists the twelve dataset categories and two machine translation systems.
The second column lists the precision results of {\toolname} under different thresholds $t$.
{\toolname} reports the suspicious issue if the distance between the translated original sentence and translated pruned sentence is larger than a threshold $t$.
For example, {\toolname} with $t=0$ \newdelete{represents}\newrevise{means} there exist \delete{words }\revise{new word occurrences} in the pruned sentence translation but not in the original sentence translation.
\revise{Similarly, }{\toolname} with $t=12$ \newdelete{represents}\newrevise{means} the number of \delete{words }\revise{new word occurrences} in the pruned sentence translation but not in \delete{the }the original sentence translation is larger than 12.
The remaining columns list the precision results of the five compared techniques.
We also present the total results calculated by the all twelve dataset categories in the middle part and bottom part of Table \ref{tab:precision}.
Each cell is represented as $x(y)$, where $x$ refers to the precision value (defined in Equation \ref{equation:precision}) and the $y$ refers to the number of all issues reported by the studied techniques.

Table \ref{tab:precision} shows when the \delete{distance }threshold \revise{value} is at its lowest (i.e., $t = 0$), {\toolname} \delete{can }\revise{is able to} report 2,140 and 2,128 erroneous issues with a 64.5\% and 65.4\% precision on average for Google Translate and Bing Microsoft Translator.
For example, when testing Google Translate with Business* dataset, {\toolname} reports 1,254 erroneous issues, while 1,094 of them contain translation errors.
We also show {\toolname} results under different \delete{distance thresholds }\revise{threshold values}, which represent the degree of inconsistency in reported issues.
In general, we will reasonably achieve less number of issues by setting a larger \delete{distance }threshold \revise{value}, but with a higher precision.
For example, we can obtain 118 erroneous issues overall for Google Translate when $t=10$, which \delete{achieve }\revise{achieves} 27\% precision improvement against $t=0$.
More importantly, we can achieve 100\% precision with Politics* dataset for the both translators when $t>4$. 

We also compare {\toolname} under different \delete{distance thresholds }\revise{threshold values} against state-of-the-art techniques.
For direct comparison, we focus on the top-1 (i.e., the translation that is most likely to contain errors) results of SIT, because SIT returns top-$k$ results for each original sentence.
In particular, the top-1 output of SIT contains (1) the original sentence and its translation and (2) the top-1 generated sentence and its translation. 
TransRepair reports a list of suspicious sentence pairs and we regard each reported pair as a suspicious issue. 

If we want to detect as many erroneous issues as possible, {\toolname} with \revise{a} lowest threshold (i.e., $t = 0$) is able to find 831 $\sim$ 3,185 erroneous issues, improving state-of-the-art techniques by at least 400\% on all datasets with competitive precision.
For example, when testing Google Translate with Business* dataset, {\toolname} \delete{can report }\revise{reports} 1,094 erroneous issues with 87.2\% precision, while SIT, PatInv, RTI, TransRepair and CAT \delete{can only report }\revise{only reports} 38, 75, 58, 74 and 168 erroneous issues with less than 80\% precision, respectively.
If we want to get more accurate results, {\toolname} \delete{can }\revise{is able to} outperform state-of-the-art techniques for both translators on average when $t>0$.
For example, compared with SIT, PatInv, RTI, TransRepair and CAT, {\toolname} with $t=6$ \delete{can improve }\revise{improves} the precision by 22.9\%, 19.9\%, 38.9\%, 17.7\% and 19.8\% when testing Google Translate, by 22.0\%, 47.2\%, 38.9\%, 27.4\%, 18.2\% when testing Bing Microsoft Translator with \revise{a} comparable amount of erroneous issues.
In Table \ref{tab:precision}, bold cells indicate the optimal precision among {\toolname} with $t=6$, SIT, PatInv, RTI, TransRepair and CAT.
In detail, {\toolname} achieves better precision performance on 22 testing scenarios, while CAT and SIT outperform on Tech and Health datasets when testing Google Translate and Bing Microsoft Translator, respectively.
We believe the results have shown the superiority of {\toolname}. 
As real-world source sentences are almost unlimited (e.g., news network), we could always set a high $t$ to get valuable erroneous issues with high precision.

\delete{\textit{False Positives and False Negatives.}}
\delete{
After analyzing the \newdelete{experiment}\newrevise{experimental} results,  there are still some false positives and false negatives in our approach.
We conclude them as follows.
First, a pruned phrase could have a different correct translation compared with the original phrase.
For example, ``the owner'' has several correct meanings (e.g., 拥有者 and 主人), while ``the owner of Carrier'' has a specific meaning (i.e., 公司拥有者) in the context ``Carrier''. 
However, we could maintain an alternative translation dictionary to alleviate this kind of false positive.
Meanwhile, a filtering mechanism can be introduced to ensure the degree of preserved context between the \delete{new}\revise{newly} generated and original sentence (in Section \ref{sec:generating-pairs}).
Second, the dependency parser that we use to parse the sentence could return wrong or inaccurate results. 
For example, the relation between ``that'' and ``employ'' is misidentified as obj for the sentence ``It is believed in the field that Amazon employs more PhD economists than any other tech company'', which leads to the generation of invalid sentences. 
Third, several generated sentences are not valid because the pruning strategy we define does not guarantee to take into account all situations in real world sentences.
For example, {\toolname} will remove the words ``of the WTO'' simultaneously because the dependency relation between ``of'' and ``WTO'' is ``case" (detailed in Table \ref{TAB:mapping}) for the sentence ``China is not the litmus test of the WTO or the world trade.''.
However, {\toolname} fails to consider \revise{the fact that} the \revise{``}case\revise{''} relation is further below the conjunction relation between the phrase ``the WTO'' and ``the world trade\delete{``}\revise{''}.
Natural language is unstructured and extremely complex in the real world, and it is extremely challenging for \delete{rules-based}\revise{rule-based} techniques (such as {\toolname}) to take all possible situations into consideration.
Despite that, {\toolname} still achieves state-of-the-art precision for testing machine translation systems.
In the future, we attempt to design more mature pruning rules for different types of sentences and analyze the impact of such rules on sentence validity.
}

\begin{table}[t]
    \footnotesize
    \caption{The number of detected translation errors}
    \label{tab:number-of-errors}
    \centering
    \begin{tabular}{l|c|c|c|c|c|c}
    \toprule
         & {\toolname} & SIT  & PatInv & RTI & TransRepair & CAT \\
    \midrule
    Google Politics & \textbf{379}   & 11    & 36    & 31    & 11    & 134 \\
    Google Business & \textbf{683}   & 15    & 3     & 46    & 18    & 124 \\
    Google Culture & \textbf{552}   & 7     & 17    & 40    & 17    & 102 \\
    Google Sport & \textbf{320}   & 17    & 217   & 42    & 42    & 84 \\
    Google Tech & \textbf{363}   & 5     & 15    & 12    & 9     & 217 \\
    Google Travel & \textbf{604}   & 3     & 10    & 32    & 14    & 189 \\
    Google Health & \textbf{305}   & 16    & 15    & 27    & 20    & 171 \\
    Google Life & \textbf{380}   & 7     & 18    & 33    & 16    & 158 \\
    Google Legal & \textbf{354}   & 13    & 15    & 32    & 7     & 157 \\
    Google Opinion & \textbf{344}   & 10    & 19    & 37    & 9     & 100 \\
    Google Politics* & \textbf{372}   & 69    & 79    & 62    & 85    & 136 \\
    Google Business* & \textbf{417}   & 57    & 76    & 51    & 75    & 173 \\
    \midrule
    Google Sum   & \textbf{5,073}  & 230   & 520   & 445   & 323   & 1,745 \\
    \midrule
    Bing Politics & \textbf{485}   & 13    & 13    & 39    & 12    & 134 \\
    Bing Business & \textbf{535}   & 17    & 6     & 41    & 18    & 130 \\
    Bing Culture & \textbf{579}   & 5     & 3     & 65    & 11    & 149 \\
    Bing Sport & \textbf{257}   & 17    & 24    & 62    & 31    & 177 \\
    Bing Tech & \textbf{388}   & 9     & 4     & 13    & 6     & 225 \\
    Bing Travel & \textbf{718}   & 5     & 9     & 26    & 12    & 169 \\
    Bing Health & \textbf{312}   & 16    & 12    & 39    & 17    & 144 \\
    Bing Life & \textbf{395}   & 12    & 7     & 49    & 25    & 130 \\
    Bing Legal & \textbf{352}   & 19    & 15    & 50    & 11    & 167 \\
    Bing Opinion & \textbf{422}   & 7     & 25    & 39    & 7     & 84 \\
    Bing Politics* & \textbf{334}   & 67    & 50    & 55    & 45    & 132 \\
    Bing Business* & \textbf{323}   & 52    & 44    & 61    & 69    & 162 \\
    \midrule
    Bing Sum & \textbf{5,100}  & 239   & 212   & 539   & 264   & 1,803 \\
    \bottomrule
    \end{tabular}
\end{table}

\subsection{Erroneous Translations}
\label{sec:errors}
We have presented that STP achieves a much better performance in terms of the number of reported issues with comparable precision.
In this section, we further analyze how many translation errors {\toolname} \delete{can }\revise{is able to} detect.
For all reported erroneous issues in Table \ref{tab:precision}, we use exact matching to remove duplicated erroneous translations.
Thus, we identify all unique translation errors in erroneous issues (i.e., the same translation error will be counted only once no matter how many times it appears).

\subsubsection{Number of Erroneous Translations}
We present the number of translation errors in Table \ref{tab:number-of-errors}.
The first column also lists the twelve dataset categories and two machine translation systems, and the remaining columns show \revise{the} number of erroneous translations detected by {\toolname} and five compared techniques, respectively.
The results of total detected erroneous translations under all the twelve dataset categories when testing Google Translate \delete{and }and Bing Microsoft Translator are also listed in the middle part and bottom part of Table \ref{tab:number-of-errors}.
The best results are shown in bold.
From Table \ref{tab:number-of-errors},  {\toolname} \delete{can detect }\revise{detects} 257$\sim$ 718 erroneous translations, which is significantly better than other techniques (3$\sim$69 for SIT, and 3$\sim$217 for PatInv, 12$\sim$65 for RTI, 7$\sim$85 for TransRepair and 84$\sim$225 for CAT).
On average, {\toolname} outperforms the most recent technique CAT by 290\% and 282\% for Google Translate \delete{and }and Bing Microsoft Translator, respectively.
It is fundamentally difficult for {\toolname} to achieve much better performance in terms of the number of reported erroneous translations with comparable precision.
We observe that STP can generate \newdelete{a mass of}\newrevise{a large number of} pruned sentences for almost all original sentences by recursively removing \delete{irrelevant }contextual words, while existing work can only test a small number of original sentences due to the limitation of adopted DL models.
If we intend to have a higher precision by setting a larger \delete{distance }threshold \revise{value}, we will reasonably obtain fewer erroneous translations.

\begin{figure}[t]
\centering
	\subfigure[Google Translate] {
		\includegraphics[width=0.38\columnwidth]{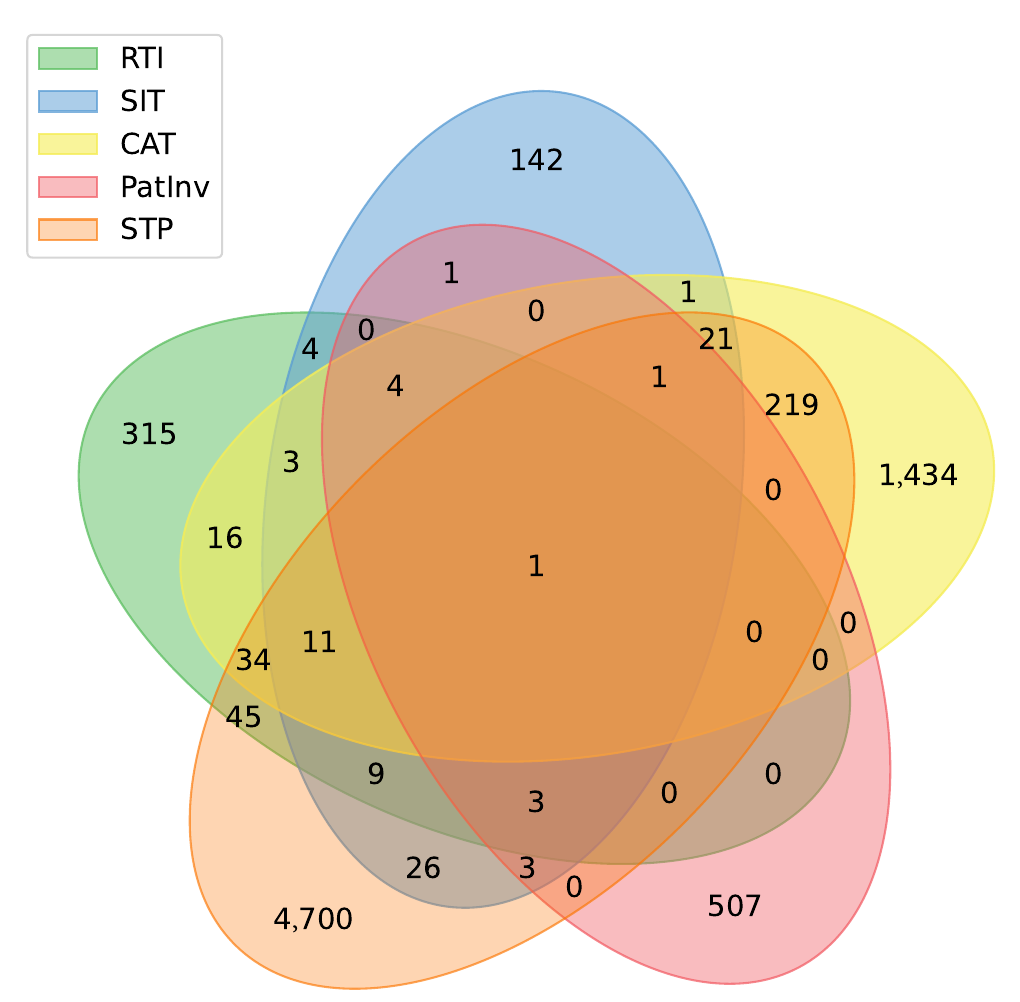}
	}
	\hspace{10mm}
	\subfigure[Bing Microsoft Translator] {
		\includegraphics[width=0.38\columnwidth]{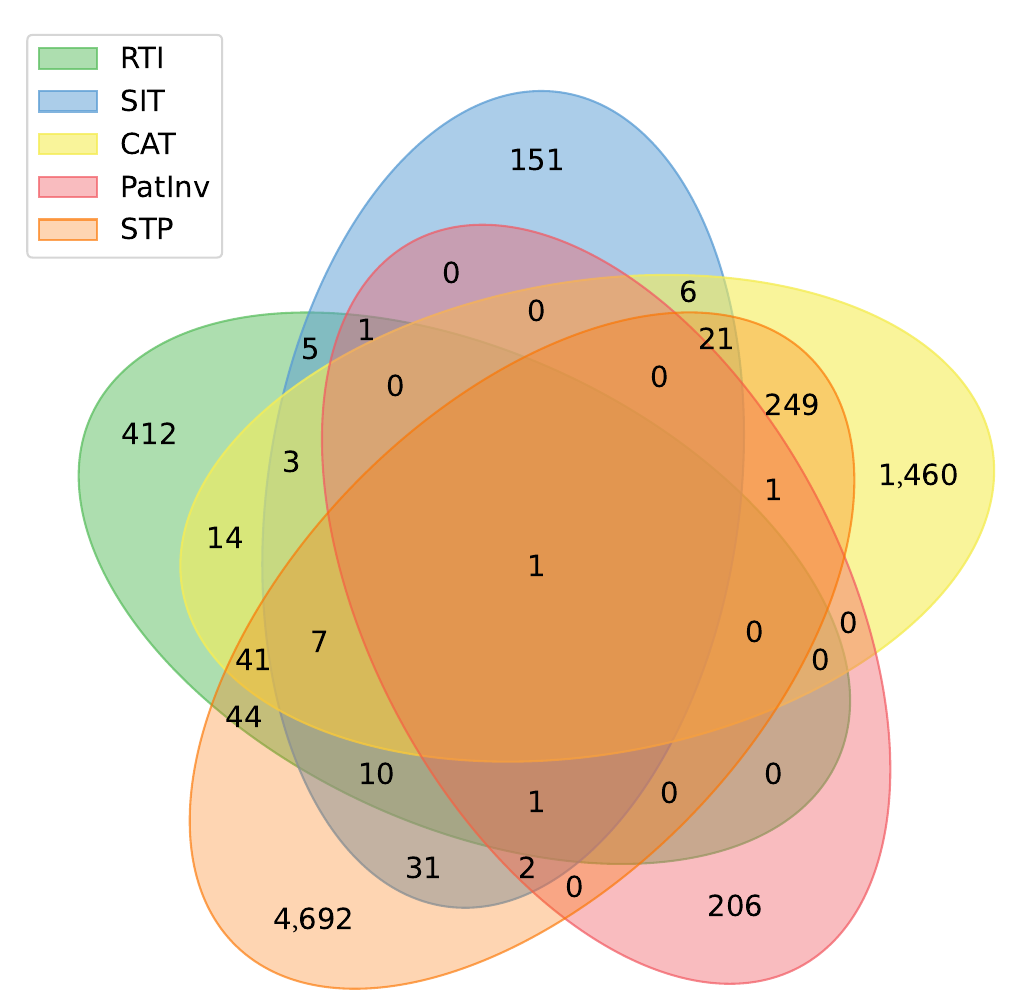}
	}
	\caption{Erroneous translations by different approaches}
	\label{fig:erroneous-translations}
\end{figure}

\subsubsection{Overlap of Erroneous Translations}
We further analyze the overlap of translation errors found by {\toolname} and other techniques.
We show {\toolname} with four best-performing compared techniques due to page limit.
Specifically, TransRepair is not included, as CAT is an extension of TransRepair and achieves better performance in terms of precision and the number of erroneous issues.
Figure \ref{fig:erroneous-translations} presents the erroneous translations reported by different approaches via Venn diagrams.

Figure~\ref{fig:erroneous-translations} shows 7,500 erroneous translations from Google Translate and 7,337 erroneous translations from Bing Microsoft Translator can be detected by the combination of all approaches, resulting in 14,837 erroneous translations in our experiment.
Particularly, there are 9,392 erroneous translations unique to {\toolname}, while there are 293, 713, 727, 2,894 erroneous translations unique to SIT, PatInv, RTI and CAT, respectively.
The improvement reaches 711.93\% on average.
After inspecting all the erroneous translations, we observe that {\toolname} is effective at reporting translation errors for \revise{both original and} pruned sentences \revise{(illustrated in Section \ref{section: a_motivating_example})}.
Meanwhile, the unique errors to RTI are mainly from the extracted phases, which have similar translations in different contexts.
The unique errors to CAT mainly come from similar sentences of one number difference (e.g., ``good'' $\rightarrow$ ``bad'').
We also find there are 219 and 249 erroneous translations \revise{that} can be detected by both {\toolname} and CAT.
After our careful analysis, the overlapped errors are mostly the original source sentences.
The high degree of overlap demonstrates that {\toolname} and the most recent technique CAT are quite effective in detecting errors for original sentences.
We will discuss it further in Section \ref{sec:num_ori}.
\revise{
For example, in Table \ref{tab:example-overlap}, we present an example that Google Translate mistranslates where ``to thread right now'' is not translated.
This erroneous translation can be detected by both {\toolname} and CAT.
{\toolname} generates a pruned sentence ``It's a very, very difficult path for banks to tread right now, Knightley said.'' by removing a word ``central'', while CAT generates a similar sentence ``It ' s a very, very difficult line for central banks to tread right now, Knightley said.'' by replacing ``path'' with ``line''.
The translation results of the two generated sentences contain the missing translation text in the original sentence.
}
Based on these results, we believe our approach complements the state-of-the-art approaches.

 \begin{table}[htbp]
     \caption{\revise{Example of overlapping erroneous translation}}
     \label{tab:example-overlap}
     \centering
     \footnotesize
     \begin{tabular}{m{2cm}|m{9cm}}
     \toprule
         \revise{Source sentence} &
         \revise{It's a very, very difficult path for central banks} \textcolor{blue}{\underline{to tread right now}}, \revise{Knightley said.} 
     \\ \midrule 
         \revise{Target sentence} & 
         \revise{奈特利说，对于中央银行来说，这是一条非常非常困难的道路。} \revise{(By Google)} \textcolor{blue}{[Erroneous Translation: ``\textit{to tread right now}" is not translated.]}
     \\ \midrule
         \revise{Target meaning} & 
         \revise{It's a very, very difficult path} \textcolor{blue}{\underline{for central banks}}, \revise{Knightley said.}
     \\ \bottomrule
     \end{tabular}
 \end{table}
 
\begin{table}[t]
  \footnotesize
  \centering
  \caption{The recall of translation errors for original sentences}
    \begin{tabular}{l|c|c|c|c|c|c}
    \toprule
          & STP   & SIT   & PatInv & RTI   & TransRepair & CAT \\
    \midrule
      Google Politics & \textbf{69\%(54)} & 6\%(5) & 3\%(2) & 19\%(15) & 5\%(4) & 50\%(39) \\
    Google Business & \textbf{66\%(53)} & 9\%(7) & 3\%(2) & 19\%(15) & 7\%(6) & 45\%(36) \\
    Google Culture & \textbf{81\%(64)} & 3\%(2) & 0\%(0) & 20\%(16) & 8\%(6) & 35\%(28) \\
    Google Sport & \textbf{75\%(41)} & 13\%(7) & 5\%(3) & 25\%(14) & 24\%(13) & 45\%(25) \\
    Google Tech & 72\%(57) & 3\%(2) & 0\%(0) & 3\%(2) & 5\%(4) & \textbf{77\%(61)} \\
    Google Travel & \textbf{91\%(49)} & 2\%(1) & 2\%(1) & 17\%(9) & 9\%(5) & 83\%(45) \\
    Google Health & 56\%(37) & 9\%(6) & 0\%(0) & 14\%(9) & 12\%(8) & \textbf{73\%(48)} \\
    Google Life & \textbf{68\%(44)} & 3\%(2) & 0\%(0) & 12\%(8) & 8\%(5) & \textbf{66\%(43)} \\
    Google Legal & 62\%(41) & 9\%(6) & 3\%(2) & 23\%(15) & 5\%(3) & \textbf{65\%(43)} \\
    Google Opinion & \textbf{69\%(35)} & 8\%(4) & 0\%(0) & 25\%(13) & 8\%(4) & 55\%(28) \\
    Google Politics* & \textbf{80\%(64)} & 45\%(36) & 4\%(3) & 28\%(22) & 4\%(3) & 45\%(36) \\
    Google Business* & \textbf{81\%(65)} & 30\%(24) & 0\%(0) & 23\%(18) & 0\%(0) & 64\%(51) \\
    \midrule
    Google Sum & \textbf{73\%(604)} & 12\%(102) & 2\%(13) & 19\%(156) & 7\%(61) & 58\%(483) \\
    \midrule
    Bing Politics & \textbf{82\%(69)} & 8\%(7) & 1\%(1) & 15\%(13) & 4\%(3) & 42\%(35) \\
    Bing Business & \textbf{87\%(69)} & 11\%(9) & 0\%(0) & 11\%(9) & 11\%(9) & 35\%(28) \\
    Bing Culture & \textbf{77\%(63)} & 0\%(0) & 0\%(0) & 22\%(18) & 4\%(3) & 49\%(40) \\
    Bing Sport & \textbf{52\%(38)} & 11\%(8) & 1\%(1) & 29\%(21) & 15\%(11) & 74\%(54) \\
    Bing Tech & 73\%(63) & 2\%(2) & 0\%(0) & 2\%(2) & 5\%(4) & \textbf{74\%(64)} \\
    Bing Travel & \textbf{80\%(64)} & 3\%(2) & 0\%(0) & 5\%(4) & 9\%(7) & 53\%(42) \\
    Bing Health & 59\%(42) & 10\%(7) & 1\%(1) & 25\%(18) & 11\%(8) & \textbf{65\%(46)} \\
    Bing Life & \textbf{85\%(58)} & 6\%(4) & 0\%(0) & 16\%(11) & 13\%(9) & 51\%(35) \\
    Bing Legal & 65\%(52) & 11\%(9) & 0\%(0) & 14\%(11) & 5\%(4) & \textbf{72\%(58)} \\
    Bing Opinion & \textbf{92\%(54)} & 3\%(2) & 0\%(0) & 19\%(11) & 5\%(3) & 32\%(19) \\
    Bing Politics* & \textbf{78\%(58)} & 42\%(31) & 3\%(2) & 28\%(21) & 3\%(2) & 41\%(30) \\
    Bing Business* & \textbf{74\%(62)} & 29\%(24) & 1\%(1) & 29\%(24) & 1\%(1) & 58\%(49) \\
    \midrule
    Bing Sum & \textbf{75\%(692)} & 11\%(105) & 1\%(6) & 18\%(163) & 7\%(64) & 54\%(500) \\
    \bottomrule
    \end{tabular}%
  \label{tab:recall}%
\end{table}%

\subsubsection{Number of Original Erroneous Translations}
\label{sec:num_ori}
Although {\toolname} \delete{can } \revise{is able to} report a large number of erroneous translations for a given sentence corpus, its performance in detecting original erroneous translations is also crucial.
For example, as shown in Section \ref{section: a_motivating_example}, if the user adopts existing machine translation testing tools to examine the news translated by Google Translate, most of the reported erroneous translations are sentences generated by the machine translation tools.
Such a low detection probability of the original sentences may limit the application of existing machine translation testing techniques in practice.

Thus, we further investigate the results of translation errors for the original sentences, which is presented in Table \ref{tab:recall}.
The first column lists the twelve dataset categories and two machine translation systems, and the remaining columns show \revise{the} number of erroneous translations for the original sentences detected by {\toolname} and five compared techniques, respectively.
Each cell is represented as $x(y)$, where $x$ is the recall value (i.e., the ratio of reported translation errors only for the original sentences over all the real translation errors) and $y$ is the number of all erroneous translations for the original source sentences.
The best results are shown in bold.
We also \delete{shows }\revise{show} the total number of original erroneous translations on \newdelete{the }all twelve dataset categories and its corresponding recall value in the middle part and bottom part of Table \ref{tab:recall} for Google translate and Bing Microsoft Translator, respectively.

From Table \ref{tab:recall}, we find {\toolname} achieves a recall of 73\% for detecting original erroneous translations when testing Google Translate, which outperforms state-of-the-art techniques by 53.4\% on average (61\% for SIT, 71\% for PatInv, 54\% for RTI, 66\% for TransRepair and 15\% for CAT).
Similar comparison performance can be observed for Bing Microsoft Translator (the improvement is 56.8\% on average).
The performance of detecting original erroneous translations is crucial in practice.
Existing works can only test a small number of original sentences, leading to a small recall of errors.
{\toolname} can generate pruned sentences with different structures for a given original sentence, and detect more original erroneous translations per category.
We believe the improvements (especially for original sentences) can make contributions to pushing machine translation testing forward in the SE community.

\subsection{Types of Reported Erroneous Translation}

\begin{table}[t]
    \caption{The number of translation errors in each category}
    \label{tab:number-of-types}
    \centering
    \footnotesize
    \begin{tabular}{l|c|c|c|c|c}
    \toprule
        & under & over & word/phrase & incorrect  & unclear \\
        & translation & translation & mistranslation & modification & logic \\
    \midrule
    Google Politics & \textbf{53\%(246)} & 10\%(48) & 13\%(61) & 10\%(46) & 13\%(61) \\
    Google Business & 28\%(255) & 13\%(122) & \textbf{34\%(312)} & 7\%(59) & 17\%(158) \\
    Google Culture & 25\%(172) & 12\%(85) & \textbf{38\%(266)} & 7\%(50) & 18\%(124) \\
    Google Sport & 20\%(89) & 4\%(19) & \textbf{45\%(200)} & 8\%(34) & 23\%(101) \\
    Google Tech & \textbf{64\%(265)} & 8\%(32) & 19\%(78) & 4\%(17) & 5\%(20) \\
    Google Travel & \textbf{67\%(472)} & 3\%(20) & 17\%(122) & 2\%(16) & 11\%(77) \\
    Google Health & \textbf{38\%(149)} & 20\%(79) & 28\%(109) & 8\%(32) & 7\%(27) \\
    Google Life & \textbf{39\%(198)} & 10\%(53) & 22\%(110) & 10\%(50) & 19\%(94) \\
    Google Legal & \textbf{61\%(230)} & 8\%(29) & 29\%(107) & 1\%(5) & 1\%(3) \\
    Google Opinion & \textbf{41\%(159)} & 34\%(132) & 10\%(39) & 11\%(42) & 3\%(13) \\
    Google Politics* & 26\%(111) & 15\%(63) & \textbf{32\%(137)} & 9\%(38) & 18\%(77) \\
    Google Business* & \textbf{38\%(210)} & 18\%(98) & 22\%(119) & 5\%(26) & 17\%(96) \\
    \midrule
    Google Sum & \textbf{41\%(2,556)} & 12\%(780) & 27\%(1,660) & 7\%(415) & 14\%(851) \\
    \midrule

    Bing Politics & \textbf{36\%(221)} & 19\%(119) & 28\%(170) & 9\%(57) & 8\%(47) \\
    Bing Business & \textbf{46\%(297)} & 16\%(102) & 21\%(136) & 3\%(17) & 15\%(100) \\
    Bing Culture & \textbf{70\%(435)} & 10\%(60) & 6\%(38) & 8\%(52) & 6\%(39) \\
    Bing Sport & 22\%(77) & 13\%(44) & \textbf{36\%(124)} & 12\%(41) & 17\%(58) \\
    Bing Tech & \textbf{68\%(307)} & 12\%(56) & 8\%(36) & 3\%(15) & 9\%(40) \\
    Bing Travel & \textbf{50\%(396)} & 9\%(68) & 23\%(183) & 5\%(39) & 14\%(112) \\
    Bing Health & \textbf{47\%(173)} & 15\%(56) & 24\%(88) & 6\%(21) & 9\%(32) \\
    Bing Life & \textbf{40\%(204)} & 4\%(22) & 30\%(152) & 7\%(38) & 18\%(94) \\
    Bing Legal & \textbf{49\%(191)} & 6\%(22) & 32\%(125) & 7\%(28) & 7\%(26) \\
    Bing Opinion & \textbf{29\%(151)} & 22\%(115) & 27\%(140) & 9\%(47) & 12\%(61) \\
    Bing Politics* & 31\%(143) & 10\%(46) & \textbf{33\%(152)} & 8\%(35) & 18\%(82) \\
    Bing Business* & 23\%(96) & 15\%(61) & \textbf{35\%(144)} & 8\%(33) & 20\%(81) \\
    \midrule
    Bing Sum & \textbf{44\%(2,691)} & 13\%(771) & 24\%(1,488) & 7\%(423) & 13\%(772) \\
    \bottomrule
    \end{tabular}
\end{table}

In this section, \revise{we evaluate the effectiveness of {\toolname} by further analyzing the diversity of detected translation errors.}
\revise{
In the literature \cite{2021He,2020He,2020Gupta}, the translation errors are mainly divided into five types: under-translation, over-translation, word/phrase mistranslation, incorrect modification, and unclear logic.
We analyze all unique translation errors detected by \newdelete{{\toolname}in}\newrevise{{\toolname} in} Section \ref{sec:errors} and classify each error into at least one type.
We report that {\toolname} is able to detect all these five types of translation errors.
}
\delete{
we report that {\toolname} can detect translation errors with various types, including under-translation, over-translation, word/phrase mistranslation, incorrect modification, and unclear logic.
}

Table \ref{tab:number-of-types} presents the number of translations that have a specific type of error.
The first column lists the twelve dataset categories and two machine translation systems, and the remaining columns list the comparison results for the five error categories.
The total results under all the twelve dataset categories are also shown in the middle part and bottom part of Table \ref{tab:number-of-types}.
The comparison result in each cell is represented as $x(y)$, where $x$ is the ratio of translation errors in this error category over the translation errors in all five error categories and $y$ is the number of erroneous translations in this category.
We can observe that Google Translate and Bing Microsoft Translator have a very similar distribution of error types overall.
For example, under-translation is the most common translation error, with over 2,500 errors for both translators.
We think the occurrence is related to our pruning  mechanism, as the removed words in the source sentence can be effectively reflected on the translation result.
Moreover, different dataset categories have diverse performances regarding the error types.
For example, the Travel dataset has more than 700 erroneous translations, while the Legal dataset has only less than 400 errors for both translators.
We think the occurrence is closely related to the training data.
This indicates there exists an imbalance issue with existing training data in different categories of datasets, which is out of the scope of our work.
We highly recommend the researchers conduct thorough evaluations for neural machine translation imbalance issues and explore how machine translation systems perform under different categories of datasets.

We \revise{also} list examples the five types of erroneous translation \revise{detected by {\toolname}} in Table \ref{tab:example-under-trabslation} to Table \ref{tab:example-unclear-logic}.
The first row shows the source sentence that is mistranslated by the translators.
The second row shows the translated sentence in the target language and explanation of the error.
The third row shows the meaning of translated sentence in the source language.
We present the types of erroneous translation in detail as follows.

\subsubsection{Under-translation}
Under-translation \delete{refers }\revise{means} that some words in the source language sentence are not translated to the target language sentence.
For example, in Table \ref{tab:example-under-trabslation},
the source sentence emphasizes that there is no limit on the number of things to borrow, while ``as much as they need'' is not translated by Google Translate\footnote{https://edition.cnn.com/2019/03/18/politics/trump-student-loan-limit-cap/index.html. Accessed August, 2022.}.
 Under-translation often leads to the lack of crucial information and different semantics meanings between the source sentence and the target sentence.

 \begin{table}[htbp]
     \caption{Example of under-translation error detected}
     \label{tab:example-under-trabslation}
     \centering
     \footnotesize
     \begin{tabular}{m{2cm}|m{9cm}}
     \toprule
         Source sentence & But parents of undergraduates and graduate students face no such limits, and can borrow \textcolor{blue}{\underline{as much as they need with the price tag}} set by schools.\\
   	\midrule 
         Target sentence& 但是，本科生和研究生的父母没有这种限制，他们可以根据学校设定的价格向他们借钱。 (By Google) \textcolor{blue}{[Erroneous Translation: ``\textit{as much as they need}" is not translated.]} \\ 
    \midrule
         Target meaning & But parents of undergraduates and graduate students face no such limits, and can borrow \textcolor{blue}{\underline{with the price}} tag set by schools. \\ 
    \bottomrule
     \end{tabular}
 \end{table}

 \subsubsection{Over-translation}
Over-translation \delete{refers }\revise{means} that some words in the target language sentence are not translated from any words in the source sentence or translated multiple times unnecessarily. 
For example, in Table \ref{tab:example-over-trabslation}, ``the almost anxiety provoking magnitude'' is translated twice by Google Translate in the target, while this phrase only appears once in the source sentence\footnote{https://edition.cnn.com/2019/03/13/tech/amazon-economists/index.html\label{example:over}. Accessed August, 2022.}.
Besides, the ``locations'' is translated into ``business locations'' by Google Translate, while ``business'' does not appear in the original sentence\textsuperscript{\ref{example:over}}.
Over-translation often brings unnecessary information in the target sentence and easily misleads people.  
 \begin{table}[htbp]
     \caption{Example of over-translation error detected}
     \label{tab:example-over-trabslation}
     \centering
     \footnotesize
     \begin{tabular}{m{2cm}|m{9cm}}
     \toprule
         Source sentence&
         You should be agonizingly jealous of the sorts of problems we work on and the almost \textcolor{blue}{\underline{anxiety provoking}} magnitude of data with which we get to work.
     \\ \midrule
         Target sentence&
         您应该非常嫉妒我们正在处理的各种问题，以及几乎令人不安的令人不安的数据量。 (By Google) \textcolor{blue}{[Erroneous Translation: ``\textit{anxiety provoking}" is translated twice.]}
     \\ \midrule
         Target meaning & 
         You should be agonizingly jealous of the sorts of problems we work on and the almost \textcolor{blue}{\underline{ anxiety provoking anxiety provoking}} magnitude of data.
     \\ \midrule \midrule
         Source sentence& 
         Imagine if you are a very large retailer that has a large number of \underline{locations}, said Martin Fleming.
      \\ \midrule 
          Target sentence& 
          马丁·弗莱明（Martin Fleming）说，想像一下，如果您是一家拥有大量营业地点的大型零售商。 (By Google) \textcolor{blue}{[Erroneous Translation: ``\textit{location}" is translated into ``营业地点'', where ``营业'' does not appear in the source sentence.]}
      \\ \midrule 
          Target meaning & 
          Imagine if you are a very large retailer that has a large number of \textcolor{blue}{\underline{business locations}}, said Martin Fleming. 
      \\ \bottomrule
     \end{tabular}
 \end{table}

\subsubsection{Word/phrase Mistranslation}
Word/phrase Mistranslation \delete{refers }\revise{means} that words or phrases in the source sentence are translated incorrectly to the target sentence.
For example, in Table \ref{tab:example-mistrabslation}, ``the elite University of California system'' is translated to ``the elite university system at the University of California''\footnote{https://edition.cnn.com/2019/03/19/politics/college-education-scandal-inequality-higher-education/index.html\label{example:mis}. Accessed August, 2022.}.
Besides, ``the rugged campaign images'' is mistranslated into ``the election images'', while the original sentence describes a perfume advertising campaign\footnote{https://edition.cnn.com/style/article/adam-driver-burberry-ltw/index.html. Accessed August, 2022.}.

 \begin{table}[htbp]
     \caption{Example of word/phrase mistranslation error detected}
     \label{tab:example-mistrabslation}
     \centering
     \footnotesize
     \begin{tabular}{m{2cm}|m{9cm}}
     \toprule
         Source sentence& 
         \textcolor{blue}{\underline{The elite University of California system}} has significantly increased its Latino enrollment.
     \\ \midrule
         Target sentence& 
         加州大学的精英大学系统极大地增加了拉丁裔的入学率。 (By Google) \textcolor{blue}{[``\textit{The elite University of California system}" is incorrectly translated into ``加州大学的精英大学系统'' which means ``\textit{The elite university system at the University of California}''.]}
     \\ \midrule 
         Target meaning & 
         \textcolor{blue}{\underline{The elite university system at the University of California}} has significantly increased Latino enrollment.
     \\ \midrule \midrule
         Source sentence& 
         The actor's fans propose that the rugged \textcolor{blue}{\underline{campaign images}} were a unanimous societal response to the question.
     \\ \midrule
         Target sentence & 
         这位演员的粉丝认为，粗犷的竞选形象是对这个问题的一致社会回应。 (By Google) \textcolor{blue}{[``\textit{campaign images}" is incorrectly translated into ``竞选形象'', and the correct translation should be ``活动图片'' in the news article.]}
     \\ \midrule 
         Target meaning & 
         The actor's fans propose that the rugged \textcolor{blue}{\underline{election images}} were a unanimous societal response to the question.
     \\ \bottomrule
     \end{tabular}
 \end{table}

\subsubsection{Incorrect Modification}
Incorrect Modification \delete{refers }\revise{means} that some \delete{modifiers }\revise{adjuncts} modify the wrong elements in the target sentence.
For example, in Table \ref{tab:example-incorrect-modification}, ``leaders'' is modified by ``who mimic his own brashness and disregard for political norms'' in the source sentence, while the clause modifies another element ``those'' in the target sentence\footnote{https://edition.cnn.com/2019/03/19/politics/donald-trump-jair-bolsonaro-brazil-white-house/index.html. Accessed August, 2022.}.

 \begin{table}[htbp]
     \caption{Example of incorrect modification error detected}
     \label{tab:example-incorrect-modification}
     \centering
     \footnotesize
     \begin{tabular}{m{2cm}|m{9cm}}
     \toprule
         Source sentence& 
         In a world of perceived foes, Trump has often looked to \textcolor{blue}{\underline{leaders}} who mimic his own brashness and disregard for political norms \textcolor{blue}{\underline{as allies}}.
     \\ \midrule 
         Target sentence & 
         在一个充满敌意的世界中，特朗普经常寻找模仿自己的傲慢并无视政治规范作为盟友的领导人。 (By Google) \textcolor{blue}{[Erroneous Translation: ``\textit{leaders}" is incorrectly modified by ``\textit{allies}''.]}
     \\ \midrule
         Target meaning & 
         In a world of perceived foes, Trump has often looked to those who mimic his own brashness and disregard for political norms \textcolor{blue}{\underline{as allies' leaders}}.
     \\ \bottomrule
     \end{tabular}
 \end{table}

\subsubsection{Unclear Logic}
Unclear Logic \delete{refers }\revise{means} that all the tokens or phrases are translated correctly but the sentence logic is not correct.
For example, in Table \ref{tab:example-unclear-logic}, ``The UC Latino students'' and ``Latinos in the community college system'' are two objects for comparison\textsuperscript{\ref{example:mis}}.
However, Google Translate does not understand the logical relation between them and thinks ``the community college system'' is the context of the sentence.
Although all words \delete{are }are correctly translated, the target sentence has a totally different meaning from the source sentence.

 \begin{table}[htbp]
     \caption{Example of unclear logic error detected}
     \label{tab:example-unclear-logic}
     \centering
     \footnotesize
     \begin{tabular}{m{2cm}|m{9cm}}
     \toprule
         Source sentence& 
         The UC Latino students are 25 times more likely to finish their degrees on time than \textcolor{blue}{\underline{Latinos in the community college system}}.
     \\ \midrule 
         Target sentence& 
         在社区大学系统中，加州大学拉丁裔学生按时完成学位的可能性是拉丁裔的25倍。 (By Google) 
         \textcolor{blue}{[Erroneous Translation: ``\textit{in the community college system}" is incorrectly translated as the context in the target sentence.]}
     \\ \midrule 
         Target meaning & 
         \textcolor{blue}{\underline{In the community college system}}, the UC Latino students are 25 times more likely to finish their degrees on time than Latinos.
     \\ \bottomrule
     \end{tabular}
 \end{table}

\subsection{Efficiency of Our Approach}
\label{sec:efficiency}
In this section, we analyze the efficiency of {\toolname} in terms of the running time.
\newrevise{Following existing studies~\cite{2020He,2021He, 2020Gupta, sun2022improving}, we run all involved techniques and report their time consumption on the same initial corpus (discussed in Section~\ref{sec:dataset}) and hardware environment (discussed in Section~\ref{sec:hardware}) to ensure a fair comparison.
It is worth noting that, unlike the fuzzing research~\cite{li2023accelerating} that usually sets the same time budget, we execute all techniques without a corresponding time limit.
The difference lies in that, fuzzing mainly focuses on test input generation, which can generate unlimited inputs to induce software systems crashes, thus often requiring a time limit to terminate execution. 
However, machine translation testing mainly relies on metamorphic testing, which designs metamorphic relations (discussed in Section~\ref{sec:generating-pairs}) and generates a limited number of test inputs to violate the relation, thus resulting a in limited running time overall.
Thus, we evaluate the running time of {\toolname} on the twelve
datasets and two translators, which is a common practice in the machine translation testing community~\cite{2020Sun}.
}
To mitigate the effect of randomness, following existing studies \cite{2020He,2021He}, we repeat {\toolname} 10 times in the same experimental setting and report the average running time.
Table \ref{tab:running-time} presents the average comparison results when testing the two translators.
The first column lists {\toolname} and five compared techniques.
The second column lists the number of generated sentences, the average time (in seconds) for each sentence generation, translation collection, and erroneous translation detection.
The remaining columns list the detailed running time of the twelve dataset categories and their average time.
We also present the average running time in the last row of each technique under the twelve dataset categories.
The best results per dataset category are in bold.
Each cell is represented as $x(y)$, where $x$ refers to the total running time per dataset category and $y$ refers to the overall running time per sentence.

From Table \ref{tab:running-time},  we can find that {\toolname} \delete{spend }\revise{spends} about 6 minutes per dataset category on average. 
Specifically, the step of collecting translation results via the translator API takes up over 60\% of the total running time. 
In our implementation, we invoke the translator API once for each source sentence and thus the network communication time is included.
Table \ref{tab:running-time} also presents the running time of state-of-the-art techniques using the same experimental settings.
We find that {\toolname} takes 149.8\% and 442.5\% more running time than RTI and TransRepair, but it can generate 392.8\% and 1886.2\% more sentences per dataset category on average.
Meanwhile, {\toolname} takes 15.8\% and 9.3\% less running time than SIT and CAT, with 9.5\% and 90.1\% more sentences than SIT and CAT.
We also find that only PatInv generate more sentence than {\toolname} by 441.0\%, but it consumes 570.7\% more running time than {\toolname}.

On average, Table \ref{tab:running-time} shows {\toolname} requires 0.39 seconds for each sentence generation, translation collection, and erroneous translation detection, improving the efficiency of state-of-the-art techniques by 22.0\%$\sim$46.6\%.
Specifically, {\toolname} consumes 0.09 seconds to generate a sentence, which is 52.6\% and 35.7\% faster than PatInv and CAT.
This improvement is attributed to the fact that, {\toolname}, unlike most techniques, does not require the employment of large-scale language models.
Among these techniques, TransRepair relies on a dictionary of similar words to directly perform word replacement, thus taking the least time cost (0.01 seconds per sentence).
The generation of the dictionary consumes more than 11 hours in our experiment setting, which is not mentioned in Table \ref{tab:running-time}.
Compared with {\toolname}, CAT directly \delete{adopt }\revise{adopts} BERT to perform word replacement and candidate filtering, resulting in more generation time (0.14 seconds per sentence).
We find {\toolname} takes about 0.02 seconds to detect erroneous translation on average, improving \delete{star-of-the-art }\revise{state-of-the-art} techniques (except PatInv) by 50\%$\sim$92.6\%, as the bag-of-words \delete{distance }\revise{metric} is very lightweight and fast in practice.
PatInv adopts a simple string comparison to compare the translations of the original and generated sentences, so PatInv performs best for erroneous translation detection (0.01 seconds on average).
Regarding translation collection, we use the same translator API to get the translation results, so the time cost is similar across different techniques.
We also find the translation time of {\toolname} is 15.2\%$\sim$49.1\% faster than other techniques.
Similar findings can be observed for RTI.
This is mainly because most of the source texts that need to be translated are phrases or pruned sentences rather than complete sentences in the dataset.
Based on the above observations, it is concluded that {\toolname} achieves comparable efficiency to state-of-the-art techniques in terms of running time.

\begin{table}[htbp]
  \centering
  \footnotesize
  \setlength\tabcolsep{2pt}
  \caption{Running time of different approaches}
    \label{tab:running-time}
    \begin{tabular}{m{0.6cm}<{\centering}|m{1.5cm}<{\centering}|cccccccccccc|c}
        \toprule
          &       & PO    & BU    & CU    & SP    & TE    & TR    & HE    & LI    & LE    & OP    & PO*   & BU*   & AVE \\
    \midrule
    \multirow{6}[4]{*}{\begin{sideways}SIT\end{sideways}} & Sentence & 405   & 817   & 578   & 1259  & 601   & 195   & 525   & 611   & 687   & 710   & 2084  & 1941  & 867.7 \\
          & Generation & 0.05  & 0.03  & 0.03  & 0.03  & 0.03  & 0.05  & 0.04  & 0.03  & 0.03  & 0.03  & 0.03  & 0.02  & 0.03 \\
          & Translation & 0.42  & 0.46  & 0.44  & 0.45  & 0.42  & 0.44  & 0.43  & 0.42  & 0.42  & 0.46  & 0.39  & 0.37  & 0.42 \\
          & Detection & 0.08  & 0.05  & 0.06  & 0.04  & 0.06  & 0.02  & 0.04  & 0.03  & 0.04  & 0.04  & 0.09  & 0.03  & 0.05 \\
\cmidrule{2-15}          & \multirow{2}[2]{*}{Total} & 259.3 & 486.7 & 349.8 & 693.2 & 342.9 & 143.4 & 311.0   & 338.6 & 374.6 & 421.2 & 1089.3 & 865.1 & 472.9 \\
          &       & (0.55) & (0.54) & (0.53) & (0.52) & (0.51) & (0.51) & (0.51) & (0.48) & (0.49) & (0.53) & (0.51) & (0.42) & (0.50) \\
    \midrule
    \multirow{6}[4]{*}{\begin{sideways}PatInv\end{sideways}} & Sentence & 3293  & 4262  & 3599  & 6972  & 3877  & 1846  & 3731  & 3347  & 4036  & 3884  & 11581 & 11251 & 5139.9 \\
          & Generation & 0.11  & 0.27  & 0.21  & 0.23  & 0.14  & 0.11  & 0.17  & 0.21  & 0.2   & 0.27  & 0.16  & 0.16  & 0.19 \\
          & Translation & 0.44  & 0.44  & 0.37  & 0.42  & 0.43  & 0.48  & 0.43  & 0.47  & 0.45  & 0.51  & 0.44  & 0.44  & 0.44 \\
          & Detection & \textbf{0.01} & \textbf{0.01} & \textbf{0.01} & \textbf{0.01} & \textbf{0.01} & \textbf{0.01} & \textbf{0.01} & \textbf{0.01} & \textbf{0.01} & \textbf{0.01} & \textbf{0.01} & \textbf{0.01} & \textbf{0.01} \\
\cmidrule{2-15}          & \multirow{2}[2]{*}{Total} & 1357  & 2743.1 & 2017.4 & 4194.8 & 1726.5 & 726.3 & 2001.9 & 2018.2 & 2069.9 & 2453.3 & 5552.5 & 5188.9 & 2670.8 \\
          &       & (0.56) & (0.72) & (0.59) & (0.66) & (0.58) & (0.60) & (0.61) & (0.69) & (0.66) & (0.79) & (0.61) & (0.61) & (0.64) \\
    \midrule
    \multirow{6}[4]{*}{\begin{sideways}RTI\end{sideways}} & Sentence & 189   & 230   & 228   & 159   & 205   & 219   & 167   & 236   & 192   & 179   & 154   & 155   & 192.7 \\
          & Generation & 0.07  & 0.07  & 0.07  & 0.07  & 0.07  & 0.06  & 0.06  & 0.06  & 0.06  & 0.07  & 0.04  & 0.05  & 0.06 \\
          & Translation & 0.35  & 0.32  & 0.30   & 0.35  & 0.32  & 0.35  & 0.31  & 0.35  & 0.37  & 0.33  & 0.28  & 0.29  & 0.33 \\
          & Detection & 0.31  & 0.26  & 0.23  & 0.27  & 0.25  & 0.27  & 0.23  & 0.29  & 0.32  & 0.29  & 0.22  & 0.24  & 0.27 \\
\cmidrule{2-15}          & \multirow{2}[2]{*}{Total} & 171.2 & 180   & 167.2 & 145.7 & 162.2 & 184.0   & 131.6 & 200.3 & 182.6 & 157.8 & 110.6 & 119.8 & 159.4 \\
          &       & (0.73) & (0.65) & (0.60) & (0.69) & (0.64) & (0.68) & (0.60) & (0.70) & (0.75) & (0.69) & (0.54) & (0.58) & (0.66) \\
    \midrule
    \multirow{6}[4]{*}{\begin{sideways}TransRepair\end{sideways}} & Sentence & 41    & 72    & 46    & 88    & 36    & 30    & 44    & 69    & 35    & 41    & 32    & 40    & 47.8 \\
          & Generation & \textbf{0.01} & \textbf{0.01} & \textbf{0.01} & \textbf{0.01} & \textbf{0.01} & \textbf{0.01} & \textbf{0.01} & \textbf{0.01} & \textbf{0.01} & \textbf{0.01} & \textbf{0.01} & \textbf{0.01} & \textbf{0.01} \\
          & Translation & 0.46  & 0.51  & 0.48  & 0.53  & 0.47  & 0.51  & 0.43  & 0.59  & 0.49  & 0.46  & 0.37  & 0.38  & 0.48 \\
          & Detection & 0.05  & 0.04  & 0.05  & 0.05  & 0.04  & 0.10   & 0.05  & 0.07  & 0.07  & 0.07  & 0.04  & 0.05  & 0.05 \\
\cmidrule{2-15}          & \multirow{2}[2]{*}{Total} & \textbf{67.0} & \textbf{90.9} & \textbf{73.3} & \textbf{104.4} & \textbf{65.2} & \textbf{68.9} & \textbf{64.0} & \textbf{104.0} & \textbf{69.1} & \textbf{68.0} & \textbf{50.2} & \textbf{55.9} & \textbf{73.4} \\
          &       & (0.52) & (0.55) & (0.54) & (0.59) & (0.52) & (0.62) & (0.49) & (0.67) & (0.57) & (0.54) & (0.42) & (0.44) & (0.54) \\
    \midrule
    \multirow{6}[4]{*}{\begin{sideways}CAT\end{sideways}} & Sentence & 500   & 500   & 500   & 500   & 500   & 500   & 500   & 500   & 500   & 500   & 498   & 500   & 499.8 \\
          & Generation & 0.15  & 0.18  & 0.16  & 0.15  & 0.14  & 0.16  & 0.13  & 0.17  & 0.16  & 0.13  & 0.09  & 0.09  & 0.14 \\
          & Translation & 0.59  & 0.59  & 0.56  & 0.58  & 0.56  & 0.6   & 0.53  & 0.59  & 0.59  & 0.51  & 0.45  & 0.46  & 0.55 \\
          & Detection & 0.04  & 0.04  & 0.07  & 0.04  & 0.03  & 0.04  & 0.03  & 0.04  & 0.04  & 0.04  & 0.03  & 0.03  & 0.04 \\
\cmidrule{2-15}          & \multirow{2}[2]{*}{Total} & 448.4 & 464.2 & 523.0   & 583.4 & 420.8 & 460.4 & 402.6 & 458.1 & 455.4 & 390.2 & 328.6 & 335.2 & 439.2 \\
          &       & (0.78) & (0.81) & (0.79) & (0.77) & (0.73) & (0.80) & (0.69) & (0.80) & (0.79) & (0.68) & (0.57) & (0.58) & (0.73) \\
    \midrule
    \multirow{6}[4]{*}{\begin{sideways}STP\end{sideways}} & Sentence & 919   & 1038  & 1155  & 821   & 852   & 1262  & 937   & 939   & 1031  & 1095  & 635   & 716   & 950.0 \\
          & Generation & 0.09  & 0.10   & 0.11  & 0.11  & 0.09  & 0.08  & 0.08  & 0.09  & 0.08  & 0.08  & 0.10   & 0.09  & 0.09 \\
          & Translation & \textbf{0.26} & \textbf{0.28} & \textbf{0.27} & \textbf{0.28} & \textbf{0.27} & \textbf{0.36} & \textbf{0.24} & \textbf{0.26} & \textbf{0.31} & \textbf{0.27} & \textbf{0.22} & \textbf{0.22} & \textbf{0.28} \\
          & Detection & 0.03  & 0.03  & 0.02  & 0.02  & 0.02  & 0.02  & 0.02  & 0.02  & 0.03  & 0.02  & 0.02  & 0.02  & 0.02 \\
\cmidrule{2-15}          & \multirow{2}[2]{*}{Total} & 371.2 & 457.8 & 484.6 & 367.6 & 350.8 & 616.2 & 351.0   & 377.6 & 465.4 & 435.5 & 238.3 & 262.9 & 398.2 \\
          &       & \textbf{(0.38)} & \textbf{(0.41)} & \textbf{(0.40)} & \textbf{(0.41)} & \textbf{(0.38)} & \textbf{(0.46)} & \textbf{(0.34)} & \textbf{(0.37)} & \textbf{(0.42)} & \textbf{(0.37)} & \textbf{(0.34)} & \textbf{(0.33)} & \textbf{(0.39)} \\
    \bottomrule
    \end{tabular}%
  \label{tab:addlabel}%
\end{table}%

\section{Discussion}
\label{sec:dis}

\subsection{Syntactic Tree Pruning}
Most existing works \cite{2020He, 2020Sun, 2020Gupta, sun2022improving} usually adopt language models to generate sentences with same structures by word-replacing (e.g., synonym replacement). 
They can only detect errors related to same structures, while \delete{fail }\revise{failing} to detect errors revealed by different structures (mentioned in Section \ref{sec:bg&mv}).
We propose the crucial semantics invariance metamorphic relation to generate sentences with different structures, which is beyond the scope of existing works.
The results demonstrate it is quite effective for {\toolname} to trigger machine translation errors with different structures, and achieve a higher recall of erroneous translations for the original sentences.
We highlight this direction to generate inputs with different structures to test machine translation systems.
Meanwhile, it is more practical to generate sentences using pre-defined operators instead of language models, as recent work shows training and deploying language models are quite costly for developers and users \cite{wang2022bridging}. 
Besides, the pre-defined operators are designed based on all available dependency types and can generate simple sentences with various structures and are generally applicable to other SE tasks, such as testing question answering systems by asking questions with different structures \cite{2020ChenQA}.

\subsection{Robust Machine Translation}
\delete{Comparing }\revise{Compared} with traditional software systems directly encoded in source code, it is more difficult to repair machine translation errors because the decision logic of neural machine translation models lies in the complex network structure and a large corpus of data.
Even if a fault-triggering test case (e.g., source language sentence) can be identified, how to automatically repair the model without introducing new errors is still a long-standing challenging task.
However, we report it is significant to use translation errors to improve machine translation systems.

\revise{
In our work, similar to most machine translation testing approaches (e.g., RTI, SIT and PatInv), {\toolname} aims to validate machine translation systems without repairing detected errors.
However, it is promising to conduct some future work on top of {\toolname} to automatically repair the detected errors and improve the robustness of the neural translation model.}
As an industrial online machine translation service, similar to traditional programming paradigms, it is easy to fix the found issue in a hard-code way without leading to negative effects.
\revise{
Besides, it is \newdelete{potential}\revise{possible} to design a post-processing strategy to repair erroneous translations for users and developers without additional manual repair efforts \cite{sun2022improving}.
}
Furthermore, it is more robust to \delete{retain }retrain or fine-tune the network with the source sentence of a translation error and its correct translation.
The reported issues may also be useful for some further debugging or repairing work (such as extracting some mistranslation patterns for a given machine translation system).

\subsection{\delete{Scalable Implementation }\revise{Scalability of {\toolname}}}
In our work, we set the source language and target language to English and Chinese, respectively, because of the knowledge background of the authors.
\revise{
Considering the fact that the two chosen languages are Top-2 most spoken languages in the world \cite{Lingua} and the English-to-Chinese setting is widely adopted in most existing machine translation testing studies \cite{2020He, 2020Sun, 2019WangWenyu, 2019Zheng, 2020Gupta}, we think this setting allows us to conduct a comprehensive comparison with selected baseline and provide reliable comparison results in our experiments.
Besides, \newdelete{The}\newrevise{the} concept of core syntactic tree methodology is general and can be built on various languages \newrevise{due to two reasons}.
First, the syntactic tree pruning is based on the basic structure and rhetorical structure theory, which is not limited to English and is a general linguistic theory across \newdelete{a mass of}\newrevise{a wide array of} mainstream languages.  
Second, the adopted Stanford Parser currently supports six languages (i.e., Arabic, Chinese, English, French, German and Spanish).}
\delete{
However, the adopted Stanford Parser currently supports six languages and the core {\toolname} property is expected to hold for most of the languages in practice.}
\revise{We also notice there exist other dependency tree parsers targeting more languages (e.g., Japanese\footnote{https://github.com/ku-nlp/bertknp. Accessed August, 2022.}). 
}
Thus, it is \revise{practical and} simple to implement {\toolname} to other languages \revise{with little engineering efforts}.
About translators, Google Translate and Microsoft Bing Translator are adopted in our work because they are widely used industrial online machine translation services and represent the state-of-the-arts.
Other popular translators (e.g., \delete{youdao }\revise{Youdao Translator} \cite{2021-Youdao}) can also be integrated into {\toolname} easily by standard translation interfaces.
Meanwhile, the method to generate new sentences (Section \ref{generate-sentences}) and detect translation errors (Section \ref{detect-errors}) is implemented as flexible \delete{components }\revise{modules} and can be enriched by other methods in the future.

\revise{The core concept of the syntactic tree methodology is also general to other NLP testing fields.
For example, similar to {\toolname}, we can generate complex sentences by context insertion to test machine translation systems based on the hypothesis that adding contextual information into a source sentence should not influence the translation results of the trunk.
It is practical to generate sentences by context insertion on top of the existing STP framework with engineering efforts. 
We can replace the context-removal-based sentence generation module with a context-insertion-based one, as other existing modules are quite suitable for the new context-insertion testing scenario. 
For example, our translation error detection module equipped with a bag-of-the-words metric is effective in finding inconsistent errors between a simple source sentence and a complex source sentence.
}

\subsection{\revise{False Positives}}
\revise{
Despite remarkable precision being achieved by {\toolname}, there are still some false positives and false negatives in our approach.
We conclude them from three main sources.
First, a pruned phrase could have a different correct translation compared with the original phrase.
For example, ``the owner'' has several correct meanings (e.g., 拥有者 and 主人), while ``the owner of Carrier'' has a specific meaning (i.e., 公司拥有者) in the context ``Carrier''. 
However, we could maintain an alternative translation dictionary to alleviate this kind of false positive.
Meanwhile, a filtering mechanism can be introduced to ensure the degree of preserved context between the newly generated and original sentence (in Section \ref{sec:generating-pairs}).
Second, the dependency parser that we use to parse the sentence could return wrong or inaccurate results. 
For example, the relation between ``that'' and ``employ'' is misidentified as obj for the sentence ``It is believed in the field that Amazon employs more PhD economists than any other tech company'', which leads to the generation of invalid sentences. 
Third, several generated sentences are not valid because the pruning strategy we define does not guarantee to take into account all situations in real-world sentences.
For example, {\toolname} will remove the words ``of the WTO'' simultaneously because the dependency relation between ``of'' and ``WTO'' is ``case" (detailed in Table \ref{TAB:mapping}) for the sentence ``China is not the litmus test of the WTO or the world trade.''.
However, {\toolname} fails to consider the fact that the ``case'' relation is further below the conjunction relation between the phrase ``the WTO'' and ``the world trade''.

To address the risks posed by the pruned sentences, we confirm that the erroneous translations labeled as true positives are syntactically and semantically correct (discussed in Section \ref{sec:labelling}).
 As a result, although a translation in a reported issue is erroneous, if its sentence in the source language is invalid, we count the translation as a false positive. 
 Besides, although we conduct a well-designed human labelling experiment, it is inevitable that a few sentences that have very small grammatical errors are not labelled by the participants. 
 In such a case, we consider the grammatical errors are typos that people introduce when typing, and mature translation systems need to handle this situation.
Natural language is unstructured and complex in the real world, and it is extremely challenging for rule-based techniques (such as {\toolname}) to take all possible situations into consideration.
\newrevise{It is promising and valuable to conducting a more in-depth analysis to investigate the impact of different rules on the testing performance.
However, considering the numerous designed rules and the extremely required manual inspection efforts in the analysis, it goes beyond the scope of our current work and can be explored in the future.}
Despite that, {\toolname} still achieves state-of-the-art precision for testing machine translation systems.
In the future, we attempt to design more mature pruning rules for different types of sentences and analyze the impact of such rules on sentence \newdelete{legality}\newrevise{validity}.
}

\revise{
\subsection{Comparison with PatInv-Remove}

In our work, following existing machine translation testing work, five state-of-the-art approaches are selected to compare against {\toolname} with 1,200 sentences from 10 major categories of news sites.
To the best of our knowledge, the selected baselines are the largest set on machine translation testing in the literature. 
As discussed in Section \ref{sec:baseline}, some baselines may have multiple variants (e.g., word-replacement based one PatInv-replace and word-removal based one PatInv-remove).
In such a case, we select the best-performing one among multiple variants according to existing relevant studies.
There may exist other possible variants that could have been used in our experiment.
For example, we select PatInv-Replace because of its superior performance against PatInv-Remove in terms of detected translation errors and precision.
However, different from selected baselines that mainly generate new test cases by replacing words, PatInv-Remove is a word-removal-based approach, which is the closest removal approach to {\toolname}.
Thus, in this section, we perform an additional evaluation by comparing {\toolname} against PatInv-Remove.

The results are presented in Table \ref{tab:removal}. 
The first column lists the twelve dataset categories and the total results calculated by all the twelve dataset categories.
The remaining columns list the two machine translation systems and two compared approaches.
Each cell is represented as $x(y/z)$, where $x$ refers to the precision value, $y$ and $z$ refer to the number of all detected and reported issues by the studied techniques.
In particular, due to page limit, we compare PatInv-Remove against {\toolname} with the threshold value of 6, which is proven to achieve remarkable precision with a comparable amount of erroneous issues in Section \ref{sec:precision}.
The detailed results are presented in our Appendix \cite{2021-Open-Datasets}.
From Table \ref{tab:removal}, we can find {\toolname} achieves 62.20\%$\sim$100.00\% precision on Google Translate, improving the metric by 6.70\%$\sim$61.90\% when compared with PatInv-Remove.
Besides, {\toolname} always detects more translation issues than PatInv-Remove (e.g., 1025 additional issues are detected on Google Translate).
When testing Bing Microsoft Translator, {\toolname} is able to improve the precision by 34.40\% while detecting 949 more translation issues in total.
\newrevise{We think the significant improvement of {\toolname} come from several differences with PatInv-Remove:
(1) PatInv-Remove assumes that sentences with different meanings should not have the same translation, while {\toolname} assumes translation results of the trunk should not influenced by eliminating contextual information from a source sentence;
(2) PatInv-Remove generates new sentences by removing a meaningful word or phrase from the sentence based on constituency trees,
while {\toolname} generates new sentences by applying a set of core semantics-preserving rules without undermining the basic structure and sentence validity based on dependency trees.
Overall, the results demonstrate that, benefiting from the novel metamorphic relation and sentence generation strategy, {\toolname} is able to perform better than PatInv-Remove in terms of both the number of detected translation issues and the reported precision.
}
}

\begin{table}[t]
  \centering
  \footnotesize
  \caption{\revise{Comparison with a state-of-the-art removal-based technique PatInv-Remove}}
    \begin{tabular}{l|c|c|c|c}
    \toprule
    \revise{} & \multicolumn{2}{c|}{\revise{Google}} & \multicolumn{2}{c}{\revise{Bing}} \\
\cmidrule{2-5}    \revise{} & \revise{STP} & \revise{PatInv-Remove} & \revise{STP} & \revise{PatInv-Remove} \\
    \midrule
    \revise{Politics} & \revise{71.4\%(35/49)} & \revise{64.7\%(33/51)} & \revise{75.5\%(37/49)} & \revise{69.0\%(20/29)} \\
    \revise{Business} & \revise{76.3\%(58/76)} & \revise{39.1\%(18/46)} & \revise{73.0\%(27/37)} & \revise{52.0\%(13/25)} \\
    \revise{Culture} & \revise{62.2\%(153/246)} & \revise{57.1\%(24/42)} & \revise{79.9\%(270/338)} & \revise{12.2\%(5/41)} \\
    \revise{Sport} & \revise{80.1\%(125/156)} & \revise{55.2\%(16/29)} & \revise{88.1\%(126/143)} & \revise{66.7\%(26/39)} \\
    \revise{Tech} & \revise{71.0\%(44/62)} & \revise{70.6\%(24/34)} & \revise{100.0\%(49/49)} & \revise{40.7\%(11/27)} \\
    \revise{Travel} & \revise{86.6\%(335/387)} & \revise{71.0\%(22/31)} & \revise{87.3\%(261/299)} & \revise{62.5\%(15/24)} \\
    \revise{Health} & \revise{84.9\%(90/106)} & \revise{39.1\%(18/46)} & \revise{64.4\%(56/87)} & \revise{47.7\%(21/44)} \\
    \revise{Life} & \revise{92.5\%(149/161)} & \revise{42.9\%(21/49)} & \revise{85.4\%(76/89)} & \revise{33.3\%(11/33)} \\
    \revise{Legal} & \revise{83.3\%(169/203)} & \revise{21.4\%(6/28)} & \revise{92.8\%(193/208)} & \revise{36.0\%(9/25)} \\
    \revise{Opinion} & \revise{85.0\%(51/60)} & \revise{54.8\%(23/42)} & \revise{73.5\%(25/34)} & \revise{60.0\%(24/40)} \\
    \revise{Politics*} & \revise{100.0\%(38/38)} & \revise{61.8\%(21/34)} & \revise{100.0\%(17/17)} & \revise{69.2\%(18/26)} \\
    \revise{Business*} & \revise{100.0\%(14/14)} & \revise{41.7\%(10/24)} & \revise{100.0\%(5/5)} & \revise{58.8\%(20/34)} \\
    \midrule
    \revise{SUM} & \revise{80.9\%(1261/1558)} & \revise{51.8\%(236/456)} & \revise{84.3\%(1142/1355)} & \revise{49.9\%(193/387)} \\
    \bottomrule
    \end{tabular}%
  \label{tab:removal}%
\end{table}%

\subsection{\delete{Random Word Deletion }\revise{Robustness of Pruned Sentences}}
\label{sec:random_deletion}

As mentioned in Section \ref{generate-sentences}, a random word detection approach would result in a low acceptability rate, while {\toolname} \delete{can }\revise{is able to} generate a large number of valid sentences based on structure syntactic relations.
We conduct a human study to check the validity of sentences generated by random work detection and STP.
As recursively removing words usually results in too many sentences, we generate 200 sentences by selecting some sentences and dropping words randomly from the dataset \cite{2021He}.
We then adopt STP to generate 200 sentences randomly from the same dataset. 
The first two authors manually check whether the generated sentences are grammatically correct and semantically reasonable.
Our manual inspection indicates that only three sentences generated by the random work deletion approach are valid, while 191 sentences generated by {\toolname} are valid.
These results show that {\toolname} \delete{can }effectively \delete{pick }\revise{picks} words to drop without breaking the validity of the original sentences.

\revise{
We notice that although massive manual efforts are devoted to analyzing all available dependency types guided by classical linguistic rhetorical structure theory, the designed pruning strategy suffers from invalid generated sentences, which is a long challenge for existing rule-based techniques \cite{2020Sikka,niklaus2019transforming}.
In the literature, a majority of existing machine translation studies adopt a word-replacement-based strategy that mainly replaces a word in the original sentence. 
In particular, they usually mask out a word and query language models (e.g., BERT) to fill the masked hole with a semantically-similar and syntactically-equivalent word (e.g., smart → cute). 
Such word-replacement-based approaches design metamorphic relations to generate sentences with the same structures, thus guaranteeing the validity of generated sentences. 
However, such same-structure-based metamorphic testing may ignore the errors revealed by sentences with different structures. 
Machine translation may return a sentence with a different structure for imperceptible perturbations due to the brittleness of the neural network. 
Thus, {\toolname} using sentences with different structures can explore behaviors of machine translation systems more fully, which is beyond the scope of existing same-structure-based works.
}

\revise{
\subsection{Actionability and Recommendation}
\label{sec:recom}
In this work, following most existing machine translation testing studies\newrevise{~\cite{2021He,sun2022improving}}, we report suspicious translation issues according to a customized threshold. 
The \newdelete{experiment}\newrevise{experimental} results reveal that {\toolname} outperforms state-of-the-art approaches on a wide range of threshold values in terms of precision and the number of reported suspicious issues, \newrevise{discussed in Section \ref{sec:precision}}.
On top of impressive results achieved by {\toolname}, we further discuss a crucial question when deploying {\toolname} to assist developers to validate machine translation systems in practice, i.e., \textit{how to choose the threshold value under different testing scenarios}?

\begin{figure}[htbp]
\centering
\graphicspath{{graphs/}}
    \includegraphics[width=0.6\linewidth]{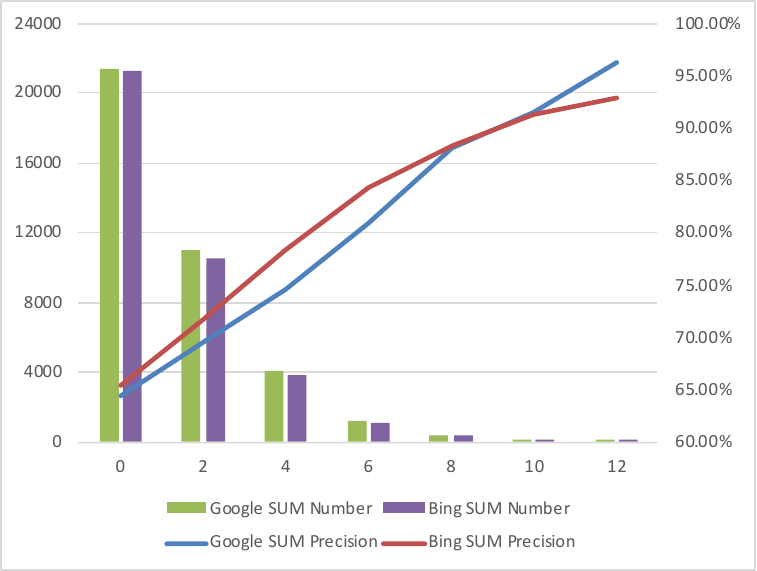}
	\caption{\newrevise{The trade-off between the precision and the number of erroneous issues.}}
	\label{fig:trade-off}
\end{figure}

\newdelete{
As discussed in Section \ref{sec:precision}, we observe a trade-off between the precision and the number of reported suspicious issues.
In particular, when we increase the threshold value, more translations are regarded as erroneous (i.e., the larger the number of issues reported), while more false positives may be introduced, resulting in a lower precision.}
\newrevise{Figure~\ref{fig:trade-off} presents the relationship between the number of detected erroneous issues and the corresponding precision under different threshold values.
We only present the overall performance of two studied translators across all datasets due to page limit.
As can be seen from Figure~\ref{fig:trade-off}, we observe a trade-off between the precision and the number of reported suspicious issues.
In particular, when we increase the threshold value, more translations are regarded as erroneous (i.e., the larger the number of issues reported), while more false positives may be introduced, resulting in a lower precision.
Similar performance can also be found in existing studies, such as RTI (Table 2 in~\cite{2021He}) and PatInv-Repalce (Table 10 in~\cite{2020Gupta}).}
Thus, we recommend different threshold values for developers to employ for testing\newdelete{, so as} to improve the actionability of {\toolname} in practice, listed as follows:
(1)	When testing resources are limited, it is obviously recommended that the highest threshold value (i.e., 12) is chosen for developers. 
In such a scenario, only inspecting fewer reported suspicious issues that have a high probability of being incorrect is more practical and reduces \newdelete{the }valuable manual effort.
(2)	When there are sufficient testing resources available, the lowest threshold value (i.e., 0) is recommended for developers, because of the more translation errors STP can detect. 
On the one hand, developers need to spend a \newdelete{mass}\newrevise{lot} of effort to assess the correctness of the reported suspicious issues manually. 
On the other hand, developers can check a considerable amount of the reported suspicious issues and reasonably find more translation errors.
(3)	When a \newdelete{well}\revise{good} trade-off between testing resources and testing effectiveness is required, the medium threshold value (i.e., 6) is recommended for developers because of \newdelete{the}\revise{its} impressive performance in terms of both precision and the number of translation errors.
In such a scenario, developers can detect more translation errors while consuming less manual inspection time than state-of-the-art approaches.
Thus, we are confident that {\toolname} is effective and easy to use for users and developers in practice.
}

\revise{
\subsection{Translation Error Detection Metrics}
In this work, we adopt a simple but effective bag-of-words strategy to calculate the distance between the original and its pruned sentence translations.
The reasons for using the bag-of-words metric are threefold.
First, the metric is suitable for our pruning design and error detection scenario.
We generate new source sentences by performing some hand-crafted word-removal operations from the original source sentence. 
In other words, we confirm that the words in the generated source sentence appear in the original sentence. 
We hope a perfect machine translation can map this relationship into the target sentences, i.e., the original target sentence contains all words that appear in the generated target sentence. 
In such a case, we adopt a bag-of-words model to denote the set difference, i.e., how many word occurrences are in the generated target sentence but not in the original target sentence. 
The experimental results also prove its effectiveness in measuring inconsistencies among translated sentences. 
Second,	the bag-of-words metric is already adopted in the literature and proves its supernormal performance in detecting translation errors. 
For example, similar to our {\toolname} finding inconsistent bugs between an original sentence and its pruned sentence, RTI aims to detect the translation errors between an original sentence and its noun phrases.
Thus, we are confident that the bag-of-words metric is reasonable to capture the inconsistent core semantics between the translations of the original and pruned sentences.
Third, the bag-of-words model is pretty lightweight and fast when deployed in practice. 
Some existing machine translation approaches tend to design complex strategies to detect translation errors (e.g., subsequence-based metric in TransRepair and structure-based metric in SIT). 
As discussed in Section~\ref{sec:efficiency}, such mechanisms usually lead to more detection time, hindering the application of such techniques in practice. 
In daily life, people rely on machine translation systems for education and communication (e.g., reading political news or articles from other countries and visiting websites with content in various languages). 
Such testing tools can help people to check the translation results in real time, which needs to be done in a very short time. 
{\toolname} adopts a bag-of-words strategy to directly calculate word occurrences with less calculation time while achieving outstanding performance.

The experimental results demonstrate the remarkable fault detection performance of the adopted bag-of-the-words when the sentence pair is made up of an original sentence and its simplified sentence \cite{2021He}.
We further explore some other metrics that consider sentence structure (e.g., constituency tree and dependency tree) from related studies, which may be more semantic-aware.
Existing studies usually generate a sentence with the same structure as the original sentence and calculate the edit distance or relation distance (e.g., the number of each phrasal type or the number of each type of dependency relations) between two similar tree structures of the translations.
However, {\toolname} generates a sentence with a different structure from the original sentence and it is improper to apply such metrics directly.
Inspired by the bag-of-words strategy that considers the word frequency in the pruned sentence translation relative to the original sentence translation, we propose a constituency tree-aware bag-of-words metric and a dependency tree-aware bag-of-words metric.
In particular, the former evaluates the distance between two sets of constituency relations by considering the set difference in the count of each phrasal type based on the intuition that the constituents of core semantics should stay the same with the original and pruned sentence where only context information differs.
Similarly, the latter evaluates the distance between two lists of dependencies by considering the set difference in the count of each type of dependency relations based on the intuition that the relationships between words that constituent core semantics will ideally remain unchanged when context information is removed.
\newrevise{We also consider the tree edit distance to detect suspicious issues. 
In particular, we parse the original translated and generated translated sentences into their constituency parse trees and then calculate the edit distance (i.e., Levenshtein distance) between the two trees. 
The tree edit distance determines how closely they match each other by calculating the minimum number of character edits (deletions, insertions, and substitutions) needed to transform one tree into the other.}

The results are presented in Table \ref{tab:metric}.
The first column lists the testing scenarios (i.e., twelve dataset categorizes $\times$ two machine translation systems).
The remaining columns list the precision results under different threshold values and evaluation metrics.
We only present the results with the threshold value as 0, 6 and 12 (recommended in Section \ref{sec:recom}) due to page limit (details in our Appendix \cite{2021-Open-Datasets}).
From Table \ref{tab:metric}, we find that the constituency tree-aware bag-of-words metric and dependency tree-aware bag-of-words metric achieve better precision performance slightly against the original bag-of-words metric.
For example, the constituency tree-aware bag-of-words metric improves the precision by 0.5\% and 1.6\% with $t=0$ when testing Google Translate and Bing Microsoft Translator.
We also find that the original bag-of-words metric detects much more erroneous issues against other metrics.
For example, the original bag-of-words metric finds 1261 erroneous issues with $t=6$ when testing Google Translate, 1126 and 695 more than other metrics.
\newrevise{Besides, we find that the tree edit metric does not achieve a better performance against our bag-of-words metric. 
The precision does not increase and the number of reported issues does not decrease when we set a larger threshold value. 
In other words, the threshold has no effect on the results when we consider the edit distance on the constituency parser tree. 
The possible reason lies in the difference in the design of our {\toolname} and existing studies. 
In particular, existing studies usually generate a sentence with the same structure as the original sentence and calculate the edit distance or relation distance (e.g., the number of each phrasal type or the number of each type of dependency relations) between two similar tree structures of the translations. However, {\toolname} generates a sentence with a different structure from the original sentence and it is improper to apply such metrics directly.}
\newrevise{
Overall, the original bag-of-words metric is effective in detecting more translation errors with considerable precision, highlighting its substantial benefit in our work.
In addition to the mentioned metrics, the results also demonstrate promising results with other tree-aware \cite{2020He,2021He} or semantic-aware metrics \cite{2020Sun,sun2022improving}, which is interesting to further explore in the future.
Besides, it is recommended to consider recent neural models (e.g., BERT) as the metrics to detect translation errors in our work.
However, whether to directly adopt existing trained models or to retrain using a large amount of labeled data, as well as how to measure the consistency of core semantics in sentence pairs, requires further exploration in the future, which goes beyond the scope of our work.
}
}

\input{tab/tab_metrics}

\section{Threats to Validity}
\label{sec:threats}

To facilitate the investigation of potential threats and to support future work, we have made the relevant materials (including source code, and translation results) available at \revise{our project website} \cite{2021-Open-Datasets}.
 Despite that, our study still faces some threats to validity, listed as follows.

The selection of the dataset might be biased.
With respect to the representativeness of the dataset, all of them are crawled from the news networks.
To mitigate the threat that the used dataset may not be representative of all real-world sentences, we adopt the dataset, which has been widely used in recent studies \cite{2020He,2021He,2020Gupta}.
Meanwhile, we randomly extract 1,000 sentences from ten different categories with varied complexity.
As such, we believe the selection strategy may not be a key point to our user study.

The second threat to validity is that the adopted dependency parser might return inaccurate even wrong dependencies.
Although, the most recent neural network-based parser made available by Stanford CoreNLP is adopted in our implementation\revise{, which parses about 100 sentences per second}.
However, the parser still has several wrong dependencies.
For example, for the sentence ``It is believed in the field that Amazon employs more PhD economists than any other tech company.'', the relation between ``that'' and ``employs'' is recognized as obj, which should be mark.
Thus, we use the Universal Dependencies as our annotation scheme, which has evolved based on the Stanford Dependencies.
Meanwhile, two authors manually check all the generated sentences to ensure the syntax and semantics legitimacy.

The \delete{last }\revise{third} threat to validity lies in the selection of baselines.
Recently, quite a few techniques have been proposed to validate machine translation systems.
We first consider all available machine translation testing techniques (i.e., \cite{2020He, 2020Sun, 2021He, 2020Gupta, 2021Ji, sun2022improving}) according to \cite{sun2022improving} when we conducted the work.
Among them, the implementation of \cite{2021Ji} is not public (last accessed in February 2022) and we fail to reproduce this work after contacting the authors. 
The remaining techniques are included in our work, and the included techniques are the same as the most recent work \cite{ sun2022improving}.
Several variants may exist for the included techniques, and we select the best one according to the existing work (shown in Section \ref{sec:baseline}).

\revise{
The final threat to validity comes from the human labelling.
In the phase of label construction, following existing studies \cite{2021Ji,2021He}, 20 \newdelete{particulars}\revise{participants} with both English and Chinese language backgrounds are responsible for deciding whether the reported issue contains a real erroneous translation.
To alleviate the influence of potential bias (e.g., imprecise labelling results) introduced by particulars, an introduction to human labelling in the form of both documents and presentation videos is given to all participants, to ensure they fully understand the experimental procedure.
To further reduce the bias when comparing different approaches, we blend all issues reported by STP and baselines to confirm the approach each issue belongs to remains unknown to the particulars.
We then guarantee that each reported issue is checked by two independent participants to reduce human impact further.
Besides, we invite participants to have an additional discussion to resolve the issues with inconsistent labels.
The high kappa scores also indicate that the bias in human labelling \newdelete{is minor to our stud}\newrevise{may not be a key point in our experiments}.
}

\section{Related Work}
\label{sec:rw}
Our work aims to validate the robustness of machine translation systems \delete{by }\revise{via} a novel metamorphic testing approach.
We divide the related work of our paper into \delete{two }\revise{three} parts: \revise{robust machine translation,} machine translation testing and metamorphic testing.

\revise{
\subsection{Robust Machine Translation}
}
 
 \revise{
 Artificial intelligence (AI) systems have been widely adopted in our daily lives thanks to the success of deep learning.
 Despite much recent research, AI systems are not as robust as we might hope sometimes.
 For example, recent research has reported a variety of adversarial examples could mislead AI systems, resulting in fatal accidents, such as autonomous cars \cite{2019Dong, 2019Zhang} and object classifiers \cite{2019Xiang, 2019Xiao}.
 Thus, a huge body of research effort has been dedicated for promoting the robustness of such AI software, focusing on testing \cite{2019Du, 2019Kim}, debugging \cite{2018Ma, 2019Wang} and training \cite{2018Lin, 2016Papernot}.
 }
 
\revise{In the NLP field, various tasks have achieved outstanding performance, such as text summarization \cite{2020Xu}, question answering \cite{2020ChenQA} and incomplete utterance rewriting \cite{2020Liu}.
 However, these NLP systems still often fail catastrophically when the given inputs are adversarially perturbed slightly, which has encouraged researchers to investigate the robustness of NLP systems. 
 For example, Jia et al. \cite{2017Jia} propose an adversarial evaluation scheme for the Stanford Question Answering Dataset (SQuAD) and find no published open-source model is robust for paragraphs that contain adversarially inserted sentences.
 Besides, Mudrakarta et al. \cite{2018Mudrakarta} analyze the robustness of three question answering models (i.e., mages, tables, and passages of text) by perturbing questions to craft a variety of adversarial examples.
 However, testing machine translation is more difficult, as one source sentence could have several correct target sentences, while the output of these systems is unique (e.g., the output of reading comprehension could be a specific person name).
 }
 
 \revise{
As a typical NLP task, machine translation aims to automatically translate text from a source language to text in a target language.
It is recently common practice to improve the robustness of machine translation software via adversarial machine learning, which aims to mislead machine translation systems with adversarial examples.
In general, the generation of these adversarial examples falls into two categories: black-box and white-box manners.
The white-box manner usually applies small perturbations that are jointly learned together with the NMT model, where complete knowledge of network structure and parameters of the machine translation model are available.
For example, Chen et al. \cite{cheng2019robust} propose a gradient-based method to craft adversarial examples guided by the translation loss of clean inputs. 
On the contrary, the black-box manner perturbs or paraphrases sentences without accessing the implementation of machine translation systems.
For example, Zhang et al. \cite{zhang2021crafting} build black-box adversarial examples based on the round-trip translation, which assumes a practical adversarial example can naturally lead to a semantics destroying round-trip translation result.
Such work can easily lead to invalid sentences, while this paper aims to generate syntactically and semantically correct test cases.
}

\subsection{Machine Translation Testing}

Machine translation testing aims to generate translation error-triggering sentences with syntactically and semantically correct \cite{2020He, 2020Sun, 2019WangWenyu, 2019Zheng, 2020Gupta, 2021He, sun2022improving}.
\revise{
Pesu et al. \cite{pesu2018monte} present the first machine translation testing approach based on metamorphic testing without human intervention or reference translation.
In particular, they design a novel metamorphic relation with the help of multiple intermediate languages, and adopt English as the source language and consider eight target languages (e.g., Chinese and Japanese).
Furthermore, they extend the above work by introducing an additional metamorphic relation, which assumes that a small change to the source sentence should not have an impact on the overall structure of the target sentence \cite{sun2018metamorphic}.}

\revise{After that, more metamorphism-based machine translation testing approaches get published in software engineering top conferences and journals, demonstrating the usefulness and potential of metamorphic testing for applications in the machine translation domain.}
\delete{Sun et al. \cite{2020Sun} combine mutation with metamorphic testing to detect inconsistency bugs and attempt to repair reported errors in a black-box or grey-box manner automatically.}
He et al. \cite{2020He} propose a novel structure-invariant testing technique \revise{(i.e., SIT)} based on the hypothesis that similar source sentences should typically exhibit similar translation results.
\revise{
In particular, they generate similar source sentences by leveraging BERT to replace one word in an original sentence with semantically-similar and syntactically-equivalent words.
They then report suspicious issues if the distance between the structure (e.g., constituency and dependency tree) of the translated original sentence and generated sentence is larger than a threshold.
Furthermore, He et al. \cite{2021He} propose a novel referential transparency test technique (i.e., RTI) based on the hypothesis that a piece of text should have similar translations in different contexts.
In particular, they extract noun phrases from the original sentence as its referential transparency by analyzing the constituent structure.
They then employ a BoW model to calculate the distance between the translation results of the original sentence and extracted phrases and report a suspicious issue if the distance is larger than a pre-defined threshold.
}
Similarly, Ji et al. \cite{2021Ji} introduce a constituency-invariant testing technique, which indicates the constituency structure of a sentence should be similar to the sentence derived from it.
\delete{We fail to} 
We do not include this work in our paper because the core implementation about sentence generation is not publicly available.
Gupta et al. \cite{2020Gupta} \delete{generate syntactically similar but semantically different sentences }\revise{propose a novel pathological invariance metamorphic testing approach (i.e., PatInv)} based on the hypothesis that the sentences of different meanings should not have the same translation.
\revise{
In particular, they generate syntactically similar but semantically different sentences by (1)replacing one word in an original sentence with a non-synonymous word via BERT (2) or removing a meaningful word or phrase from an original sentence via its constituency structure.
They then suspect an erroneous translation if the translation results of the original and generated sentences are the same via a simple string comparison.
}
Recently, Cao et al. \cite{2020Cao} propose a semantic-based machine translation testing approach (\revise{i.e., }SemMT) \delete{by transforming the sentences into regular expressions }\revise{ based on the hypothesis that regular expressions can effectively extract the semantics concerning logical relations and quantifiers in sentences}.
\revise{
In particular, they first conduct the round-trip translation to collect the intermediate and translated sentences and transform the sentences into regular expressions with existing tools.
They then capture semantic similarities over regular expressions by a set of regex-related metrics and report suspicious issues if the similarity is higher than a predefined threshold.}
However, instead of arbitrary sentence \revise{semantics}, SemMT focuses on the semantics of quantifiers and logical relations, which is out of the scope of our work.
\revise{
Sun et al. \cite{2020Sun} propose a novel approach to test machine translation (i.e., TransRepair) by combining mutation with metamorphic testing to detect inconsistency bugs and attempt to automatically repair reported errors in a black-box or grey-box manner.
In particular, they conduct context-similar word replacement to generate a mutated sentence and assume that translation results from both the original sentence and its mutant should have a certain degree of consistency modulo the replaced word.}
\delete{
Further, Sun et al. \cite{sun2022improving} propose a novel approach to test machine translation by calculating context-aware semantic similarity to identify an isotopic replacement.}
\revise{
Furthermore, Sun et al. \cite{sun2022improving} propose a novel word-replacement-based approach (i.e., CAT) to test machine translation on the top of TransRepair based on the insight that controlling the semantic difference between the replaced words is crucial in the impact of word replacement.
In particular, they employ BERT to encode the sentence context during replacement and calculate context-aware semantic similarity to identify an isotopic replacement on the top of TransRepair.}

The key idea of {\toolname} is conceptually different from most existing approaches \cite{2020He, 2020Gupta,2020Sun}, which replace a word or extract a phrase to generate new sentences.
In contrast, {\toolname} assumes that the translation of a source sentence should be similar to the newly generated sentence generated by eliminating \delete{irrelevant context }\revise{contextual} information.

\subsection{Metamorphic Testing}

Metamorphic testing \cite{2020Chen, 2018Chen, 2016Segura} is a well-known technique to check the functional correctness of various systems in the absence of an ideal oracle.
Its key idea is to detect violations of domain-specific metamorphic relations among the inputs and outputs of multiple executions of the program under test. 
As an effective method to alleviate the oracle problem, metamorphic testing has seen successful applications in a variety of domains, including compilers \cite{2014Le, 2015Lidbury}, database systems \cite{2015Lindvall} and scientific libraries \cite{2014Zhang}.
In addition, the effectiveness of metamorphic testing has also been shown repeatedly in AI systems testing because of its ability to test non-testable applications, such as search engines \cite{2015Zhou, 2012Zhou}, autonomous driving systems \cite{2018Tian, 2018Zhang} and classifiers \cite{2008Murphy, 2011Xie}.
For example, Chen et al. \cite{2021Chen} propose three novel metamorphic relations for testing question answering software by checking its behaviors on multiple recursively asked questions that are related to the same knowledge.
\revise{
To further eliminate false positives, Shen et al. \cite{shen2022natural} propose a precise metamorphic testing approach for question answering softwares, involving five sentence-level metamorphic relations (e.g.,  inserting a redundant sentence as a clause of the original question).
}
\revise{
Yu et al. propose \cite{yu2022automated} the first metamorphic approach to test image captioning systems, which attempt to generate a brief depiction of the salient objects in an image.
They assume that the object names should exhibit directional changes after object insertion and design two metamorphic relations (i.e., object appearance and singular-plural form).
Besides, Huang et al. \cite{huang2022aeon} propose a novel approach to evaluate the NLP test cases generated by  metamorphic testing approaches based on similarity consistency and language naturalness.
}
{\toolname} is a novel metamorphic testing approach for machine translation software based on the insight that the new sentence generated by eliminating \delete{irrelevant }\revise{contextual} information should retain the core semantics of the original sentence.

\section{Conclusion}
\label{sec:conclusion}
In this paper, we present a widely applicable methodology, syntactic tree pruning, to detect erroneous translations for machine translation systems.
In contrast to existing approaches, which rely on perturbing a word or extracting specific phrases (i.e., noun phrases) in natural sentences, {\toolname} assumes the pruned sentence should retain the core semantics of the original sentence.
In particular, given an arbitrary sentence, {\toolname} generates new sentences via a novel core semantics-preserving pruning strategy on the syntactic tree level,
and then pairs the generated and original sentence by the designed metamorphic relation.
{\toolname} reports suspicious issues if the translation results break the consistency property in semantics via a bag-of-words model.
\delete{Because of }\revise{Benefitting from} the distinct concept, {\toolname} has reported a diversity of erroneous translations, most of which can not be found by existing approaches.
As a result, {\toolname} successfully detect 5,073 translation errors for Google Translate and 5,100 translation errors for Microsoft Translator, respectively, with comparable precision (i.e., 64.5\% and 65.4\%).
{\toolname} also achieve a recall of 74\% when detecting erroneous translation for the 1,200 original sentences, improving state-of-the-art techniques by 55.1\% on average.

In the future, we will generalize the core concept of {\toolname} to other implementation scenarios, such as inserting an unimportant word into a source sentence without changing crucial information.
Meanwhile, it would be interesting to conduct a series of research on automated program repair for machine translation systems with the reported translation results.

\section*{Acknowledgment}
\revise{The authors would like to thank the anonymous reviewers for their insightful comments.}
This work is supported partially by the National Natural Science Foundation of China (61932012, 62141215, 62372228) and the China Scholarship Council (202306190117).

\bibliographystyle{ACM-Reference-Format}
\bibliography{reference}

\end{CJK*}
\end{document}

%% file: tab/tab_metrics.tex
\begin{sidewaystable}[htbp]
  \centering
  \scriptsize
  \caption{\revise{Comparison results with different metrics}}
  \setlength{\tabcolsep}{0.5pt}
  
    \begin{tabular}{c|cccc|cccc|cccc}
    \toprule
          & \multicolumn{4}{c|}{t=0}      & \multicolumn{4}{c|}{t=6}      & \multicolumn{4}{c}{t=12} \\
    \midrule
    Category & STP   & \newrevise{STP\_edit} & STP\_con & STP\_dep & STP   & \newrevise{STP\_edit} & STP\_con & STP\_dep & STP   & \newrevise{STP\_edit} & STP\_con & STP\_dep \\
    \midrule
    GO PO & 61.0\%(1410/2310) & \newrevise{57.6\%(1717/2982)} & \textbf{66.1\%(926/1400)} & 62.8\%(1113/1772) & \textbf{71.4\%(35/49)} & \newrevise{60.2\%(1571/2610)} & 11.1\%(1/9) & 56.5\%(13/23) & N.A.  & \newrevise{\textbf{63.7\%(1362/2137)}} & N.A.  & N.A. \\
    GO BU & 86.1\%(2895/3364) & \newrevise{85.8\%(3369/3925)} & 85.1\%(1825/2145) & \textbf{85.4\%(2319/2715)} & 76.3\%(58/76) & \newrevise{86.1\%(3095/3595)} & \textbf{100.0\%(17/17)} & 88.1\%(52/59) & \textbf{100.0\%(6/6)} & \newrevise{86.0\%(2643/3072)} & \textbf{100.0\%(6/6)} & \textbf{100.0\%(6/6)} \\
    GO CU & \textbf{60.5\%(2574/4255)} & \newrevise{59.1\%(2882/4879)} & 59.1\%(1657/2805) & 59.6\%(2021/3393) & 62.2\%(153/246) & \newrevise{60.4\%(2688/4454)} & \textbf{84.6\%(11/13)} & 80.4\%(45/56) & N.A.  & \newrevise{\textbf{61.9\%(2378/3841)}} & N.A.  & N.A. \\
    GO SP & 55.7\%(1315/2361) & \newrevise{53.7\%(1480/2758)} & 54.5\%(806/1479) & \textbf{56.3\%(1017/1807)} & \textbf{80.1\%(125/156)} & \newrevise{55.0\%(1386/2522)} & 70.6\%(12/17) & 78.7\%(37/47) & \textbf{100.0\%(21/21)} & \newrevise{57.0\%(1236/2167)} & \textbf{100.0\%(3/3)} & \textbf{100.0\%(3/3)} \\
    GO TE & 63.1\%(1283/2033) & \newrevise{57.3\%(1525/2663)} & 63.5\%(805/1267) & \textbf{66.0\%(1065/1614)} & 71.0\%(44/62) & \newrevise{59.8\%(1414/2363)} & \textbf{75.0\%(12/16)} & 67.3\%(35/52) & N.A.  & \newrevise{\textbf{62.3\%(1241/1993)}} & N.A.  & N.A. \\
    GO TR & 67.2\%(3087/4591) & \newrevise{65.4\%(3453/5278)} & \textbf{67.3\%(2012/2988)} & 66.8\%(2513/3762) & 86.6\%(335/387) & \newrevise{66.7\%(3225/4833)} & \textbf{93.1\%(27/29)} & 78.9\%(116/147) & 95.5\%(21/22) & \newrevise{68.7\%(2886/4199)} & N.A.  & \textbf{100.0\%(5/5)} \\
    GO HE & 51.0\%(1579/3099) & \newrevise{48.4\%(1727/3569)} & 51.5\%(1056/2051) & \textbf{52.1\%(1332/2557)} & 84.9\%(90/106) & \newrevise{49.8\%(1617/3247)} & 75.0\%(3/4) & \textbf{87.3\%(48/55)} & N.A.  & \newrevise{\textbf{52.6\%(1453/2764)}} & N.A.  & N.A. \\
    GO LI & 55.8\%(1837/3295) & \newrevise{53.7\%(2013/3746)} & 56.6\%(1319/2330) & \textbf{58.4\%(1525/2613)} & \textbf{92.5\%(149/161)} & \newrevise{55.3\%(1892/3424)} & 90.9\%(20/22) & 89.0\%(81/91) & N.A.  & \newrevise{\textbf{57.1\%(1722/3017)}} & N.A.  & N.A. \\
    GO LE & 62.2\%(1848/2971) & \newrevise{56.7\%(2010/3544)} & 63.6\%(1120/1761) & \textbf{63.7\%(1569/2463)} & \textbf{83.3\%(169/203)} & \newrevise{59.0\%(1893/3210)} & \textbf{83.3\%(10/12)} & 76.0\%(57/75) & \textbf{66.7\%(2/3)} & \newrevise{62.7\%(1718/2738)} & N.A.  & N.A. \\
    GO OP & 61.1\%(1553/2541) & \newrevise{58.2\%(1797/3088)} & \textbf{64.3\%(1048/1630)} & 61.6\%(1204/1953) & 85.0\%(51/60) & \newrevise{60.2\%(1660/2757)} & \textbf{100.0\%(18/18)} & 88.5\%(54/61) & N.A.  & \newrevise{64.1\%(1442/2251)} & N.A.  & \textbf{100.0\%(1/1)} \\
    GO PO* & 82.5\%(933/1131) & \newrevise{79.2\%(1072/1353)} & 82.5\%(570/691) & \textbf{83.3\%(707/849)} & \textbf{100.0\%(38/38)} & \newrevise{82.4\%(985/1195)} & \textbf{100.0\%(3/3)} & 81.8\%(18/22) & N.A.  & \newrevise{\textbf{86.7\%(800/923)}} & N.A.  & N.A. \\
    GO BU* & 87.2\%(1094/1254) & \newrevise{84.4\%(1310/1553)} & \textbf{89.1\%(770/864)} & 88.6\%(906/1022) & \textbf{100.0\%(14/14)} & \newrevise{86.6\%(1180/1363)} & \textbf{100.0\%(1/1)} & \textbf{100.0\%(10/10)} & N.A.  & \newrevise{\textbf{89.1\%(994/1116)}} & N.A.  & N.A. \\
    \midrule
    \multirow{2}[2]{*}{GO SUM} & 64.50\% & \newrevise{61.90\%} & 65.00\% & \textbf{65.20\%} & 80.90\% & \newrevise{63.50\%} & \textbf{83.90\%} & 81.10\% & 96.20\% & \newrevise{65.80\%} & \textbf{100.00\%} & \textbf{100.00\%} \\
          & (21408/33205) & \newrevise{(24355/39338)} & (13914/21411) & \textbf{(17291/26520)} & (1261/1558) & \newrevise{(22606/35573)} & \textbf{(135/161)} & (566/698) & (50/52) & \newrevise{(19875/30218)} & \textbf{(9/9)} & \textbf{(15/15)} \\
    \midrule
    BI PO & \textbf{72.9\%(1685/2310)} & \newrevise{69.1\%(2060/2982)} & 71.1\%(996/1400) & 70.9\%(1257/1772) & \textbf{75.5\%(37/49)} & \newrevise{70.9\%(1851/2610)} & 11.1\%(1/9) & 60.9\%(14/23) & N.A.  & \newrevise{\textbf{73.8\%(1578/2137)}} & N.A.  & N.A. \\
    BI BU & 66.6\%(2129/3198) & \newrevise{62.5\%(2460/3933)} & \textbf{71.3\%(1364/1913)} & 67.3\%(1700/2525) & 73.0\%(27/37) & \newrevise{63.9\%(2282/3571)} & \textbf{100.0\%(6/6)} & 62.2\%(28/45) & N.A.  & \newrevise{\textbf{64.6\%(1965/3043)}} & N.A.  & N.A. \\
    BI CU & 75.1\%(3127/4166) & \newrevise{70.9\%(3464/4885)} & \textbf{77.2\%(2295/2971)} & 74.8\%(2604/3482) & 79.9\%(270/338) & \newrevise{73.5\%(3289/4473)} & \textbf{96.6\%(85/88)} & 88.6\%(209/236) & \textbf{88.9\%(8/9)} & \newrevise{76.9\%(2980/3877)} & N.A.  & 0.0\%(0/2) \\
    BI SP & 51.5\%(1197/2326) & \newrevise{47.8\%(1314/2751)} & 55.4\%(730/1317) & \textbf{55.5\%(926/1668)} & \textbf{88.1\%(126/143)} & \newrevise{50.3\%(1244/2471)} & 62.5\%(5/8) & 78.7\%(37/47) & \textbf{100.0\%(6/6)} & \newrevise{53.3\%(1107/2078)} & N.A.  & N.A. \\
    BI TE & 64.8\%(1338/2064) & \newrevise{60.7\%(1615/2662)} & 64.3\%(868/1349) & \textbf{66.9\%(1112/1661)} & \textbf{100.0\%(49/49)} & \newrevise{62.9\%(1494/2377)} & 80.0\%(12/15) & 90.3\%(65/72) & N.A.  & \newrevise{\textbf{65.1\%(1293/1986)}} & N.A.  & N.A. \\
    BI TR & 69.9\%(3185/4558) & \newrevise{69.1\%(3651/5283)} & \textbf{73.6\%(2263/3074)} & 72.1\%(2720/3771) & 87.3\%(261/299) & \newrevise{70.2\%(3389/4829)} & \textbf{92.7\%(38/41)} & 91.2\%(196/215) & \textbf{100.0\%(2/2)} & \newrevise{71.7\%(2994/4178)} & \textbf{100.0\%(1/1)} & \textbf{100.0\%(8/8)} \\
    BI HE & \textbf{51.1\%(1550/3033)} & \newrevise{48.6\%(1738/3573)} & 50.0\%(1047/2093) & 50.5\%(1288/2548) & 64.4\%(56/87) & \newrevise{50.4\%(1620/3213)} & \textbf{71.4\%(10/14)} & 86.4\%(38/44) & N.A.  & \newrevise{53.3\%(1429/2681)} & N.A.  & \textbf{100.0\%(3/3)} \\
    BI LI & 66.6\%(2124/3190) & \newrevise{63.3\%(2360/3731)} & \textbf{69.6\%(1486/2136)} & 68.0\%(1725/2536) & 85.4\%(76/89) & \newrevise{65.5\%(2222/3391)} & \textbf{100.0\%(3/3)} & \textbf{100.0\%(11/11)} & N.A.  & \newrevise{\textbf{68.9\%(1995/2896)}} & N.A.  & N.A. \\
    BI LE & \textbf{60.7\%(1750/2881)} & \newrevise{56.1\%(1992/3551)} & 60.0\%(940/1567) & 58.2\%(1305/2242) & \textbf{92.8\%(193/208)} & \newrevise{57.8\%(1844/3189)} & 88.8\%(71/80) & 84.4\%(92/109) & 92.2\%(47/51) & \newrevise{61.2\%(1662/2715)} & 84.6\%(22/26) & \textbf{93.8\%(61/65)} \\
    BI OP & 60.1\%(1536/2556) & \newrevise{59.1\%(1824/3084)} & 60.5\%(1059/1750) & \textbf{61.1\%(1203/1970)} & \textbf{73.5\%(25/34)} & \newrevise{59.7\%(1641/2747)} & 35.7\%(5/14) & 48.9\%(23/47) & \textbf{100.0\%(1/1)} & \newrevise{60.5\%(1367/2259)} & N.A.  & 28.6\%(2/7) \\
    BI PO* & \textbf{80.7\%(837/1037)} & \newrevise{76.7\%(1041/1357)} & 80.2\%(542/676) & 78.7\%(661/840) & \textbf{100.0\%(17/17)} & \newrevise{78.2\%(938/1199)} & N.A.  & 80.0\%(4/5) & N.A.  & \newrevise{\textbf{80.2\%(736/918)}} & N.A.  & N.A. \\
    BI BU* & \textbf{67.7\%(831/1227)} & \newrevise{64.9\%(1005/1548)} & 64.9\%(519/800) & 64.2\%(623/970) & \textbf{100.0\%(5/5)} & \newrevise{65.9\%(889/1348)} & 85.7\%(6/7) & \textbf{100.0\%(18/18)} & N.A.  & \newrevise{\textbf{66.4\%(700/1054)}} & N.A.  & N.A. \\
    \midrule
    \multirow{2}[2]{*}{BI SUM} & 65.40\% & \newrevise{62.30\%} & \textbf{67.00\%} & 65.90\% & 84.30\% & \newrevise{64.10\%} & \textbf{84.90\%} & 84.30\% & \textbf{92.80\%} & \newrevise{66.40\%} & 85.20\% & 87.10\% \\
          & (21289/32546) & \newrevise{(24524/39340)} & \textbf{(14109/21046)} & (17124/25985) & (1142/1355) & \newrevise{(22703/35418)} & \textbf{(242/285)} & (735/872) & \textbf{(64/69)} & \newrevise{(19806/29822)} & (23/27) & (74/85) \\
    \bottomrule
    \end{tabular}%

  \label{tab:metric}%
\end{sidewaystable}%